\documentclass[journal]{IEEEtran}
\usepackage{amsmath,amsfonts}
\usepackage{algorithmic}
\usepackage{algorithm}
\usepackage{array}
\usepackage{textcomp}
\usepackage{stfloats}
\usepackage{url}
\usepackage{verbatim}
\usepackage{graphicx}
\usepackage{cite}
\usepackage{soul}
\usepackage{color}
\usepackage{multirow}
\usepackage{subcaption}
\usepackage{pgfplots, pgfplotstable}
\usepackage{tikz}
\usepgfplotslibrary{groupplots}
\pgfplotsset{compat=newest}
\usetikzlibrary{shapes, arrows, calc, patterns, matrix, positioning,fit,decorations.markings}
\begin{document}
\title{Synthesizing Annotated Image and Video Data Using a Rendering-Based Pipeline for Improved License Plate Recognition}
\author{Andreas Spruck,~\IEEEmembership{Student Member,~IEEE,} Maximilane Gruber,~\IEEEmembership{Student Member,~IEEE,} Anatol Maier,~\IEEEmembership{Student Member,~IEEE,} Denise Moussa, J\"urgen Seiler,~\IEEEmembership{Senior Member,~IEEE,} Christian Riess,~\IEEEmembership{Senior Member,~IEEE,} and Andr\'e Kaup,~\IEEEmembership{Fellow,~IEEE}%}%
% <-this % stops a space
\thanks{Manuscript received ...}
\thanks{Andreas Spruck, Maximilane Gruber, J\"urgen Seiler, and Andr\'e Kaup are with the Chair of Multimedia Communications an Signal Processing, Friedrich-Alexander-Universit\"at Erlangen-N\"urnberg, Cauerstr.~7, 91058~Erlangen.}
\thanks{Anatol Maier, Denise Moussa, and Christian Riess~are with the Chair of Computer Science 1 (IT Security Infrastructures), Friedrich-Alexander-Universit\"at Erlangen-N\"urnberg, Martensstr.~3, 91058~Erlangen.}
\thanks{We gratefully acknowledge support by the German Federal Ministry of Education and Research (BMBF) under Grant No.~13N15319. }
}
\maketitle
\begin{abstract}
An insufficient number of training samples is a common problem in neural network applications. While data augmentation methods require at least a minimum number of samples, we propose a novel, rendering-based pipeline for synthesizing annotated data sets. Our method does not modify existing samples but synthesizes entirely new samples. The proposed rendering-based pipeline is capable of generating and annotating synthetic and partly-real image and video data in a fully automatic procedure. Moreover, the pipeline can aid the acquisition of real data. The proposed pipeline is based on a rendering process. This process generates synthetic data. Partly-real data bring the synthetic sequences closer to reality by incorporating real cameras during the acquisition process. The benefits of the proposed data generation pipeline, especially for machine learning scenarios with limited available training data, are demonstrated by an extensive experimental validation in the context of automatic license plate recognition. The experiments demonstrate a significant reduction of the character error rate  and miss rate from 73.74\% and 100\% to 14.11\% and 41.27\% respectively, compared to an OCR algorithm trained on a real data set solely. These improvements are achieved by training the algorithm on synthesized data solely. When additionally incorporating real data, the error rates can be decreased further. Thereby, the character error rate and miss rate can be reduced to 11.90\% and 39.88\% respectively. All data used during the experiments as well as the  proposed rendering-based pipeline for the automated data generation is made publicly available under~\textit{(URL will be revealed upon publication)}. 
\end{abstract}
\section{Introduction}
\IEEEPARstart{T}{he} recent advances in the field of neural networks (NN) rapidly changed the state of the art and made NN applications the new standard technique in almost every image and video processing application. NNs are commonly applied whenever the content of an image or scene should be interpreted. Since the NN AlexNet~\cite{Krizhevsky2012} first won the ImageNet Large Scale Visual Recognition Challenge~(ILSVRC)~\cite{Russakovsky2015} in 2012 while clearly outperforming all competitors, much research was conducted in the area of machine learning and NNs. Due to the development of increasingly efficient and robust algorithms combined with the availability of powerful hardware, especially GPUs, impressive results could be achieved. NNs learn to extract suitable features for solving the current task by themselves during a training procedure. This enables NNs to achieve high performance. During the training procedure, samples are presented to the NN that hold the answer to the current task in an accompanying label. During the training, the network solves the current task based on the parameters learnt up to then. The result proposed by the NN is then compared to the ground-truth annotation. Based on the outcome of the comparison the weights of the network are updated such that the network output approaches the labels closer. The learnt weights within the NN steer the feature extraction process. \par
The performance of the trained NN largely depends on the used training data set. Due to the importance of the training data sets, a large variety of different data sets was proposed during the last years for many different application scenarios. Nevertheless, the availability of suitable training data sets is still a limiting factor for many NN applications, as the generation of such data sets is very elaborate. Data sets used for the training of NNs usually require several hundreds of thousands annotated samples in order to detect and extract reasonable features from the samples. Already the acquisition of that many samples is very elaborate. However, the labeling of all these samples is even more time consuming and thereby costly, as this task is commonly executed manually. Even though the labeling is sometimes aided by crowd working platforms as Amazon Mechanical Turk (AMT) or similar to speed up the process~\cite{Russakovsky2015}. \par 
An application with restricted availability of suitable training data is the recognition of license plates. Here, not only the elaborate collection and preparation process hinders the development of suitable databases, but also restrictions regarding data protection and privacy limit the collection and publication of databases for many countries. This impairs the development of algorithms for the Automatic License Plate Recognition (ALPR), even though the demand for ALPR systems increases with the further growing availability of camera data in the traffic and transportation sector. The field of application is widespread reaching from civil applications such as toll collection~\cite{Du2012}, payment of parking fees~\cite{Anagnostopoulos2014}, access control~\cite{Anagnostopoulos2010}, and the analysis of traffic volume~\cite{Du2012} to forensic applications~\cite{Anagnostopoulos2006}. Recordings acquired with mobile devices play an important and increasing role within forensic investigations, as with the unbroken popularity of smartphones, video recording devices are available almost any time and everywhere. For the recognition of license plates a manifold of algorithms have been proposed within the recent years \cite{Anagnostopoulos2008,Du2012,Silva2018,Chan2020,DeOliveira2021}. \par 
The recent developments in the field of object detection and Optical Character Recognition (OCR) also influenced the ALPR algorithms. Existing approaches often combine the detection of a license plate within an image in a first step, followed by the recognition of the text on the license plate in a second step~\cite{Anagnostopoulos2008}. Most commonly these two steps are performed separately by different algorithms. \par 
With the success of NNs during the past years, nowadays, also for ALPR tasks NN approaches are very common. With the use of NNs, the recognition rate could be boosted significantly \cite{Hill2016,Silva2018}. By using NNs, it is even possible to recognize licenses plates that are not readable for human viewers \cite{Lorch2019}. \par 
NNs are nowadays most commonly applied for both of the aforementioned steps - detection and recognition. A very crucial requirement for all NN approaches is the training data. The annotation of training data in the ALPR context has to contain information regarding the license plate detection as well as recognition. For the training of the detector, the labels contain a bounding box specifying the position of the object within the current sample. To train the recognition application, a text annotation of the license plate is required. During the training phase the algorithm identifies features that are extracted from the images. Therefore, it is important that the data used in the final application is similar to the training data. Otherwise, the identified features do not produce meaningful outputs. As the license plate detector should only extract the position of the license plate within the current sample, also training data holding license plates from other countries than the desired one can be used. However, the dimensions and the general design have to be similar. The license plate recognition on the other hand is often sensitive towards the font used on the license plate. Also the used alphabet and text length might vary for license plates originating from different countries. \par 
In this paper the proposed rendering-based pipeline for the generation of annotated image and video data is used to produce ALPR data sets. In the following, the workflow of the pipeline will be demonstrated synthesizing samples of German license plates. Subsequently, the recognition performance of an algorithm trained on the synthesized data is examined to evaluate the effectiveness of the proposed pipeline. \par 
The upcoming section first summarizes related work. Especially, existing data sets and ALPR algorithms are presented in further detail. In Section~\ref{sec:pipeline}, we present the proposed rendering-based pipeline for synthesizing annotated image and video data sets. The proposed pipeline generates three different data types -- synthetic, partly-real, and real data. The generation of synthetic data is introduced in Section~\ref{sec:syn_data}, before the procedure of capturing the partly-real data is described in Section~\ref{sec:pr_data} and real data in Section~\ref{sec:real_data}. The proposed pipeline for synthesizing data is validated for a challenging ALPR scenario. The conducted experimental setup is described in Section~\ref{sec:exp_setup}. The obtained results are discussed in Section~\ref{sec:eval}. Section~\ref{sec:conclusion} summarizes and concludes this paper. 
\section{State of the Art}
Our proposed rendering-based data generation pipeline is used in this paper to improve the recognition of license plates. As ALPR is sensitive towards the used font, alphabet, and text length that varies for different countries, it is important to train the recognition algorithm on license plate samples from the same region as used during the final application. This caused the collection of a manifold of different license plate data sets. With the large variety of different data sets, different perspectives, resolutions and environmental conditions are available. The most common data sets show Brazilian~\cite{Goncalves2016,Laroca2018}, Taiwanese~\cite{Hsu2012,Hsu2017}, Chinese~\cite{Xu2018,Zhang2020} or Czech~\cite{Svoboda2016,Spanhel2017} license plates. With respect to German license plates only one publicly available data set exits. This data set was published in~\cite{Chan2020}. However, it does not contain German license plates solely but is mixed with French license plates. \par 
Table~\ref{tab:sota_datasets_sizes} gives an overview over the most common data sets related to ALPR. Most of these data sets holding annotated license plate data are too small to train large NNs, as this requires several hundreds of thousand training samples. The largest data sets listed in Table~\ref{tab:sota_datasets_sizes} hold between 140,000 and 250,000 samples of Czech or Chinese license plates, respectively. Moreover, common data sets hold mostly license plates of good visual quality, as the camera acquiring the license plate samples is oriented directly towards the license plates. For forensic ALPR applications this does not match the common use case, as here often license plates in the background of an image are of special interest. Those samples are often very small, suffer from bad illumination and show an overall bad visual quality. Therefore, existing license plate data sets often lead to insufficient results in forensic ALPR applications. \par
\begin{figure}[t]
	\centering
	\includegraphics[width=0.9\columnwidth]{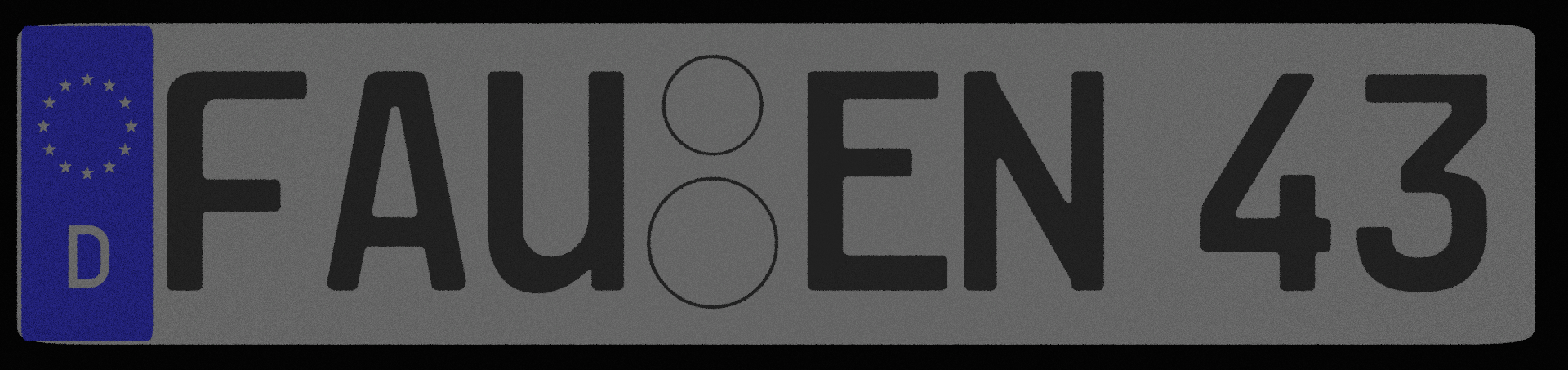}
	\caption{Example visualizing the layout of a German license plate. The first text block specifies the region where the car is registered. The second text block holds one to two arbitrary characters. The last text block contains one to four digits. The second and last text block build the individual identifier.}
	\label{fig:LP_layout}
\end{figure}
Regarding German license plates the only available data set holds 18,672 samples, where only a fraction is actually German~\cite{Chan2020}.  German license plates can be divided into three different text blocks. An example visualizing the layout of a German license plate can be seen in Figure~\ref{fig:LP_layout}. The first text block is one to three characters long and specifies the city or region where the car is registered. The second text block, located in the middle of the license plate, holds one to two arbitrary characters. The last text block is one to four digits long and may only contain numbers. However, it is specified that the overall number of characters and numbers may not exceed nine symbols~\cite{Justiz2011}. Each German license plate is unique and the personal information of the car owner may be deduced from it. Therefore, license plates are considered personal data and protected by the laws of data privacy and protection. \par 
\begin{table}
\centering
\caption{Overview over the most common data sets related to ALPR} 
\label{tab:sota_datasets_sizes}
\begin{tabular}{|m{0.25\columnwidth}<{\centering}|m{0.25\columnwidth}<{\centering}|m{0.25\columnwidth}<{\centering}|}
\hline
Data Set & Region & Number of bounding boxes \\ \hline
SIGG \cite{Goncalves2016} & Brazilian &  {2,000} \\ \hline
UFPR-ALPR \cite{Laroca2018} & Brazilian & {4,500} \\ \hline
AOLP \cite{Hsu2012} & Taiwanese & 2,049\\ \hline
AOLPE \cite{Hsu2017} & Taiwanese & 4,210\\ \hline
ReId \cite{Spanhel2017} & Czech & {182,336} \\ \hline
Svoboda et al. \cite{Svoboda2016} & Czech & 140,000 \\ \hline
CCPD \cite{Xu2018} & Chinese & {250,000} \\ \hline
CLPD \cite{Zhang2020} & Chinese & 1,200\\ \hline
TLPD \cite{Chan2020} & German/ French & 18,672 \\ \hline
CD-HARD \cite{Silva2018} & Various & 102 \\ \hline
Iranis \cite{Tourani2021} & Iranian & 83,000 \\ \hline
UCSD-Stills \cite{Dlagnekov} & US & 1169 \\ \hline
PKU Vehicle Data set \cite{Yuan2016} & Chinese & 3,977 \\ \hline
\end{tabular}
\end{table}
%
% Weiter ausführen, allgemeinen Fall besser darlegen? Was macht meine Pipeline besser? -> PR data
The acquisition of real-world data sets is very elaborate and  therefore expensive. %In the case of license plates it also impairs conflicts regarding data privacy. 
Due to this the idea of generating synthetic data sets was brought up~\cite{Hittmeir2019}. The idea of synthetic training data is also incorporated in other fields of application such as healthcare~\cite{Dahmen2019}, face analysis~\cite{Wood2021}, crowd counting~\cite{Wang2019,Wang2020}, object detection and pose estimation~\cite{Tremblay2018,Tremblay2018a,Wrenninge2018}, segmentation~\cite{Wrenninge2018,Sankaranarayanan2018}, scene understanding~\cite{Roberts2021}, or text spotting~\cite{Jaderberg2014,Gupta2016}. The benefits of synthetic data are that data can be generated and labeled automatically in large amounts without affecting data protection and privacy regulations. \par 
There are also first works examining the usage of synthetic training data for the recognition of license plates. However, these first attempts produce data with only limited realism. In~\cite{Silva2018}, synthetic data is used to train the OCR application of an ALPR system. The synthetic data are produced by plotting text strings of a fixed length of seven characters in the license plate specific font onto a textured background. Afterwards, a random geometric transformation is applied to the generated sample. In~\cite{Silvano2020}, another approach of generating synthetic data is pursued. Silvano et. al.~\cite{Silvano2020} place synthetically generated images of license plates onto existing images of cars. Thereby, they extend the number of different license plate samples without acquiring new data. Therefore, artificial images of license plates with random text labels are generated. Subsequently, the bounding box of the license plate is estimated within the existing images of cars, using a Darknet53 network~\cite{Redmon2016}. The synthetic license plate is then geometrically transformed such that its shape matches the estimated bounding box. In a final step, the transformed synthetic license plate is blended into the original image covering the original license plate. Even though this procedure is able to extend existing data sets, the produced images often contain inconsistencies, as the illumination, brightness, or blurriness of the license plate do not match with the remaining image. Therefore, the image statistics deviate strongly for different parts of the image. \par 
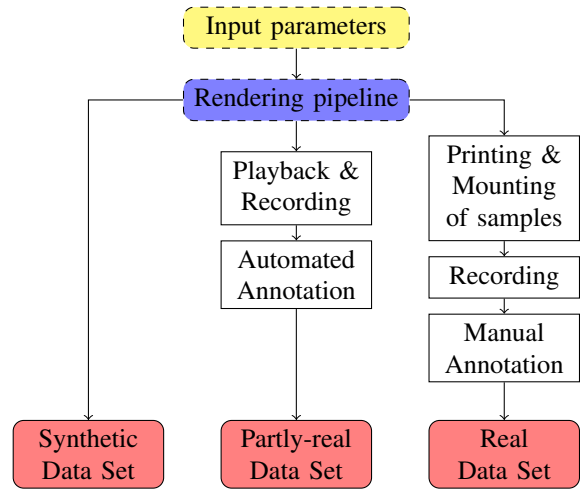
\begin{figure}[t]
	\centering
	\begin{tikzpicture}
	% Boxen
	\node[draw,dashed,fill=yellow,fill opacity=0.5,text opacity=1,minimum width=3cm,rounded corners] at (0,0) (Input) {Input parameters}; % at (6,8)  ,minimum width=12cm, minimum height=2cm
	%	\node[align=right, anchor=north east, below =0cm of csv.south east] {csv-file};
	
	\node[draw,dashed,fill=blue,fill opacity=0.5,text opacity=1,minimum width=3cm,rounded corners, below=0.4cm of Input] (Rendering) {Rendering pipeline}; % at (6,8)  ,minimum width=5cm, minimum height=2cm	
	
	%	\node[draw, align = center,rounded corners, below left=0.75cm of Rendering] (Syn_dataset) {Synthetic Data Set;\\Bounding Box \& \\ Text annotation};
	%	\node[draw, align = center,rounded corners, minimum width=1cm,below left=0.75cm of Rendering] (Syn_dataset) {Synthetic\\Data Set};
	
	\node[draw, align = center,minimum width=2cm, below=0.4cm of Rendering] (Playback) {Playback \&\\Recording};	
	\node[draw, align = center, minimum width=2cm,below=0.2cm of Playback] (Aut_Ann) {Automated\\Annotation};	
	%	\node[draw, align = center,rounded corners, below=0.75cm of Aut_Ann] (PR_dataset) {Partly-real Data Set;\\Bounding Box \& \\ Text annotation};
	%	\node[draw, align = center,rounded corners, minimum width=1cm,below=0.75cm of Aut_Ann] (PR_dataset) {Partly-real\\Data Set};

	\node[draw, align = center,minimum width=2cm, right=0.75cm of Playback] (Printing) {Printing \& \\Mounting\\of samples};	
	\node[draw, align = center, minimum width=2cm,below=0.2cm of Printing] (Rec) {Recording};	
	\node[draw, align = center, minimum width=2cm,below=0.2cm of Rec] (Man_Ann) {Manual\\Annotation};

	\node[draw, align = center,rounded corners,fill=red,fill opacity=0.5,text opacity=1, minimum width=2cm, below=0.5cm of Man_Ann] (Real_dataset) {Real\\Data Set};
	\node[draw, align = center,rounded corners,fill=red,fill opacity=0.5,text opacity=1, minimum width=2cm,left=0.75cm of Real_dataset] (PR_dataset) {Partly-real\\Data Set};
	\node[draw, align = center,rounded corners,fill=red,fill opacity=0.5,text opacity=1, minimum width=2cm,left=0.75cm of PR_dataset] (Syn_dataset) {Synthetic\\Data Set};
	
	% Pfeile
	%	\draw [->] (Light) |- (Blender.east);
	%	\draw [->] (Scene_params) |- (Blender.west);
	%	\draw [->] (String) -- (Blender);
	%	
	%	\draw [->] (Blender) -- (Render);
	%	
	%	\draw [->] (Render) -- (Syn_dataset);
	\draw[->] (Input) -- (Rendering);
	
	\draw[->] (Rendering) -- (Playback);
	\draw[->] (Playback) -- (Aut_Ann);
	\draw[->] (Aut_Ann) -- (PR_dataset);
	
	\draw[->] (Rendering.east) -| (Printing.north);
	\draw[->] (Printing) -- (Rec);
	\draw[->] (Rec) -- (Man_Ann);
	\draw[->] (Man_Ann) -- (Real_dataset);
	
	\draw[->] (Rendering.west) -| (Syn_dataset.north);
	
\end{tikzpicture}
	\caption{Overview over the proposed pipeline for the automated generation of image and video data. With the proposed pipeline, synthetic and partly real data can be generated fully automatic. The generation of real data can as well be aided by the proposed pipeline.}
	\label{fig:overview_pipeline} 
\end{figure}
To address the problem of insufficient training data, we propose a rendering-based pipeline that is capable of synthesizing new artificial data. With the proposed pipeline, data sets consisting of synthetic, partly-real and real data can be produced. The positive effects of incorporating a rendering-based pipeline were already demonstrated for a data augmentation application in the context of OCR in~\cite{Spruck2021}. The framework presented in~\cite{Spruck2021} is capable of extending existing data sets by realistic samples acquired under novel viewing angles. The rendering-based pipeline presented here does not extend existing data sets, but is syntheses entirely new data. \par 
While license plates with visually good quality can be recognized well by existing algorithms, the recognition of degraded samples is still an open issue. Even though there are several methods available that address the problem of impaired visual quality \cite{Spanhel2017,Agarwal2017,Lorch2019}, limited availability of annotated samples with visually impaired quality often prohibits development of further applications. The development of ALPR algorithms for European and especially German license plates is often restricted by missing large-scale data sets. The cause for the lack of suitable data sets are in many cases limitations due to data protection and privacy. This often prevents the recording of license plates in unconstrained public environments. Our novel rendering-based pipeline for synthesizing image and video data sets addresses this problem. In this article we focus on the problem of recognizing license plates of impaired visual quality. This scenario is especially important for forensic and criminalistic applications. The proposed pipeline can be configured to match the individual users needs. Thereby, the desired visual output quality can be steered, regarding viewing angle, illumination, and size of the license plate. All data sets synthesized by our pipeline comply to current laws regarding data protection and privacy, as the generated license plates cannot be linked to any personal data. In contrast to existing frameworks generating synthetic data, our proposed rendering-based pipeline is furthermore able to include real-world cameras into the data generation process. Thereby, the acquisition of partly-real and real data sets is enabled. The partly-real data can be generated in large amounts, as the whole procedure is automatized. Moreover, the proposed pipeline is not limited to single frames but is able to generate video sequences incorporating motion and changes in illumination. Even though the pipeline is presented for the generation of license plate data here, it is not limited to this application. It can rather be extended to any arbitrary object, by simply replacing the license plate object. \par 
\section{Proposed Rendering-based Pipeline}
\label{sec:pipeline}
Common data augmentation methods require at least a minimum number of annotated samples that are altered. Thereby the overall number of samples available during training is increased. Therefore, our proposed rendering-based pipeline is not to be confused with a data augmentation method, as it does not augment samples but synthesizes entirely new samples. Our proposed method is capable of producing large-scale image and video data sets. It generates three different types of data -- synthetic, partly-real, and real data. The whole pipeline and the individual processing steps are depicted in Figure~\ref{fig:overview_pipeline}. The basis of the proposed method builds the rendering pipeline. \par 
Synthetic data can be generated in large amounts with accompanying labels in a fully automated manner. This holds for single images as well as video sequences. The process of generating synthetic data is introduced in further detail in Section~\ref{sec:syn_data}. For the acquisition of partly-real data, synthetically generated sequences are displayed on a screen and recorded with real-world cameras. The sequences are annotated automatically in a post-processing step. %As the recording and subsequent annotation procedure of the partly-real data is more elaborate than the synthetic data, this part of the data set is smaller than the fully synthetic part. 
The procedure of generating partly-real data is further explained in Section~\ref{sec:pr_data}. The third data type are real data, where samples are captured under real-world conditions. The labeling of this type of data has to be done manually, as no previous knowledge can be used. The process of obtaining real data is explained in further detail in Section~\ref{sec:real_data}. \par %
\begin{figure}[t]
	\centering
	\resizebox{0.9\columnwidth}{!}{
		\begin{tikzpicture}
	% Boxen
	\node[draw,dashed,fill=yellow,fill opacity=0.5,rounded corners ,minimum width=9cm, minimum height=1.6cm] at (-0,0) (csv) {}; % at (6,8) 
%	\node[align=right, anchor=north east, below =0cm of csv.south east] {csv-file};
	
	\node[align=center] at (0,0) (String) {License Plate\\String}; % at (3,8) 
	\node[align=center,left= 0.2cm of String] (Scene_params) {Desired resolution,\\ desired scene length}; %  anchor=north west,
	\node[align=center, right= 0.2cm of String] (Light) {Camera, light, and\\motion trajectory\\parameters};
	
	\node[draw,dashed,fill=blue,fill opacity=0.5,rounded corners ,minimum width=5cm, minimum height=1.7cm, below=0.7cm of String] (pipeline) {}; % at (6,8) 
%	\node[align=left, anchor=west, above right =0cm of pipeline.south east] {Rendering-based\\Pipeline};
	\node[draw, below=0.9cm of String] (Blender) {3D Scene generation}; %0.7cm 
	\node[draw, align = center, below=0.2cm of Blender] (Render) {Rendering};

	\node[draw, align = center, below=0.5cm of Render] (Syn_dataset) {Synthetic Data Set;\\Bounding Box \& \\ Text annotation};
	
	% Pfeile
	\draw [->] (Light) |- (Blender.east);
	\draw [->] (Scene_params) |- (Blender.west);
	\draw [->] (String) -- (Blender);
	
	\draw [->] (Blender) -- (Render);
	
	\draw [->] (Render) -- (Syn_dataset);
	
\end{tikzpicture}
	}
	\caption{Fully automated pipeline for generating and annotating synthetic data.}
	\label{fig:syn_pipeline} 
\end{figure}
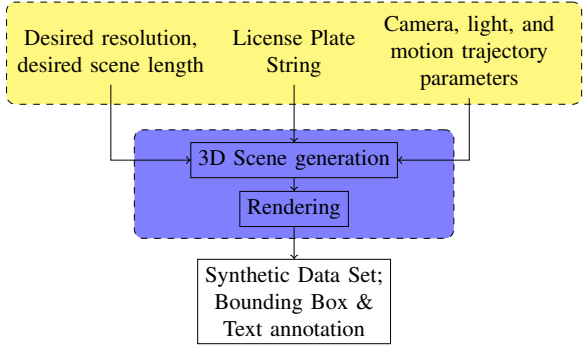
The rendering-based pipeline for synthesizing data is demonstrated for the generation of German license plates here. Throughout this work we focused on producing challenging recognition scenarios to train an algorithm that can recognize visually impaired samples.  For the implementation of the rendering pipeline the long-term supported version~2.83 of the open source software Blender~\cite{BlenderFondation2020} is used. \par
The upcoming sections present the three different data types and their generation process in detail. 
\subsection{Synthetic Data Generation}
\label{sec:syn_data}
\begin{figure}[t]
	\centering
	\includegraphics[width=\columnwidth]{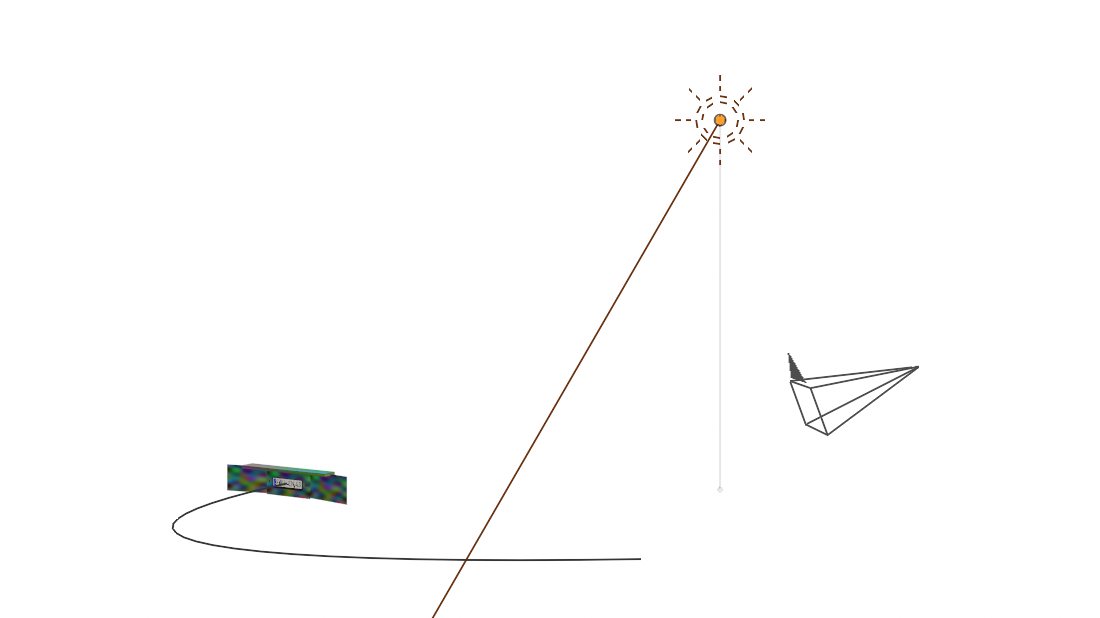}
	\caption{Example of a Blender scene generated with the proposed pipeline. The scene holds the license plate, a schematic car back, the motion path, a light source and a camera.}
	\label{fig:blender_scene}
\end{figure}
The synthetic data are produced fully automated by the proposed pipeline as depicted in Figure~\ref{fig:syn_pipeline}. The pipeline deploys the open source rendering software Blender to built a 3D~scene and render it. The software is steered entirely via Python scripts making user inputs during the data generation obsolete. The parameters configuring the scene are read from a configuration file by the scripts. The scene and framework have a modular structure allowing to switch the single components of the scene on or off. Additionally, they can be replaced individually. \par %
Blender bears the possibility to represent each component of the scene in its original size. This ensures that the scale of all components matches the real world. Moreover, the size and distances among the single objects relates to real-world scenarios. The main component of each scene is the object that should be included into the data set. Here we use German license plates as these objects. The structure and appearance of the license plates and its text is exactly specified by German authorities in \cite{Justiz2011}. Besides all distances, labels, and size also the exact font and its font size is specified. The font required for civil license plates is called FE-font and is specified in~\cite{Justiz2011} as well. Several open source implementations of this font are available. In this work, the implementation from \cite{Hoppe} is used. As each license plate is uniquely linked to the owner of the vehicle, license plates are personal data that fall under the laws of data protection. The proposed framework takes care of privacy protection by randomly generating arbitrary license plate configurations. Hence, each license plate configuration is equally probable and the framework complies with the rules of European data protection laws. \par 
By exchanging the font and position of the text boxes the pipeline can easily be configured to produce license plates from other countries as well. These changes have to be applied only once to the template. \par
Every generated scene consists of several elements that are the same in all cases. These elements are directly imported into the 3D scene during its generation. For the example of synthesizing license plate data, one of the shared elements is a schematic car rear painted with colored lowpass noise. By using the colored lowpass noise, overfitting onto the car and its color is reduced. Furthermore, a black mounting frame is part of every scene as well as the white background of the license plate and the blue country sign as specified in \cite{Justiz2011}. For data sets holding other objects than license plates, only these components of the scene have to be replaced by the desired object. The subsequent steps remain unchanged for the generation procedure, as well as the labeling process. \par 
Furthermore, each scene contains a light source, a camera, and a motion trajectory along which the object moves through the scene. Figure~\ref{fig:blender_scene} shows an exemplary scene as used for the generation of the license plate samples. This scene contains all of the above described components. The schematic car rear with the mounted license plate moves along the motion trajectory that is depicted as black curve going through the center of the license plate. On the right side of Figure~\ref{fig:blender_scene} the camera can be seen. In the top of the figure, the light source is depicted as orange circle. The main direction of illumination is marked by the orange line originating from the light source. Blender bears the possibility to individually configure the parameters of the cameras and light sources. Here, the camera is loaded from a predefined set and is randomly chosen during the generation of the configuration file. For the light sources a predefined set is available as well. This set holds configurations from a broad variety of different common light sources. The lights differ in brightness, position and beamwidth. They model common light sources such as spot lights, area lights, or sunlight. Similar holds for the cameras. The predefined cameras vary, e.g., in the position of the camera within the scene, the camera tilt, its sensor size, aperture, and focal length. The implemented cameras range from dash cams over mobile phones and security cameras to full frame cameras. Additionally, the user of the framework can configure own cameras and lights matching the individual requirements of the considered scenario. The data generation pipeline can furthermore be extended to include not only one but several light sources within the same scene. Thereby, even more complex illumination scenarios can be realized. \par 
A configuration file that holds all necessary parameters for the rendering procedure of each sequence serves as input of the generation pipeline. It is denoted by the yellow box in Figure~\ref{fig:syn_pipeline}. The configuration files are generated automatically. For the generation of license plates they hold a randomly chosen text label, as well as a camera and light source that are chosen randomly from the predefined sets. Moreover, the file contains a seed for the random number generator producing the motion trajectory. Additionally, the desired output resolution and scene length are specified. Thereby, all random choices within the synthesized data sets are made during the preparation of the configuration file. The rendering-based pipeline itself is deterministic. This ensures the reproducibility of all generated sequences. \par 
For the motion trajectory contained in the scene, two points are randomly chosen within the field of view of the current camera. For this random choice the seed specified in the configuration file is used. The trajectory between these two chosen points is then interpolated by a spline interpolation. \par  %
The rendering-based pipeline, denoted by the blue box in Figure~\ref{fig:syn_pipeline} generates frame-wise loss-lessly coded sequences in png format. Besides the image data, the framework outputs annotation data as well. This annotation data is stored in an .xml-file for each sequence individually. The file holds all hyper parameters controlling the render environment such as the used render engine and the render resolution. Moreover, parameters defining the scene itself, e.g. used light and camera, are saved. Beside this global scene information, the annotation of the current object is saved. For the license plate data, the text label and bounding box of the license plate is stored for every frame. Furthermore, the annotation file contains flags that indicate whether the object in the current frame is occluded and if so whether the occlusion is on the left or right side or on the top or bottom of the object. Occlusions may occur when parts of the object move out of the field of view of the current camera. 
\subsection{Partly-real Data Generation}
\label{sec:pr_data}
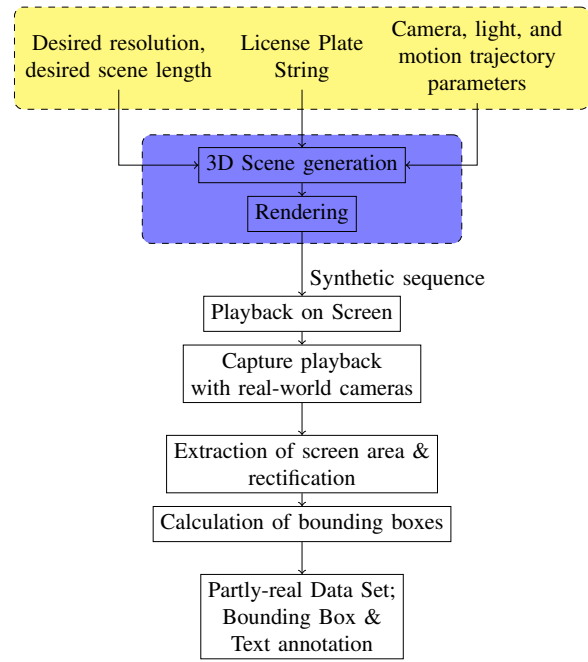
\begin{figure}[t]
	\centering
	\resizebox{0.9\columnwidth}{!}{
		\begin{tikzpicture}
	% Boxen
	\node[draw,dashed,fill=yellow,fill opacity=0.5,rounded corners ,minimum width=9cm, minimum height=1.6cm] at (-0,0) (csv) {}; % at (6,8) 
%	\node[align=right, anchor=north east, below =0cm of csv.south east] {csv-file};
	
	\node[align=center] at (0,0) (String) {License Plate\\String}; % at (3,8) 
	\node[align=center,left= 0.2cm of String] (Scene_params) {Desired resolution,\\ desired scene length}; %  anchor=north west,
	\node[align=center, right= 0.2cm of String] (Light) {Camera, light, and\\motion trajectory\\parameters};
	
	\node[draw,dashed,fill=blue,fill opacity=0.5,rounded corners ,minimum width=5cm, minimum height=1.7cm, below=0.7cm of String] (pipeline) {}; % at (6,8) 
%	\node[align=left, anchor=west, above right =0cm of pipeline.south east] {Rendering-based\\Pipeline};
	\node[draw, below=0.9cm of String] (Blender) {3D Scene generation}; %0.7cm 
	\node[draw, align = center, below=0.2cm of Blender] (Render) {Rendering};
	
	\node[draw, align = center, below=1cm of Render] (Display) {Playback on Screen};
	\node[draw, align = center, below=0.2cm of Display] (Capture) {Capture playback\\with real-world cameras};
	
	\node[draw, align = center, below=0.5cm of Capture] (Rectifiataion) {Extraction of screen area \&\\rectification};
	\node[draw, align = center, below=0.2cm of Rectifiataion] (bbox_calc) {Calculation of bounding boxes};
	
	\node[draw, align = center, below=0.5cm of bbox_calc] (pr_dataset) {Partly-real Data Set;\\Bounding Box \&\\Text annotation};
	
	% Pfeile
	\draw [->] (Light) |- (Blender.east);
	\draw [->] (Scene_params) |- (Blender.west);
	\draw [->] (String) -- (Blender);
	
	\draw [->] (Blender) -- (Render);
	
	\draw [->] (Render) -- (Display) node[near end, anchor=west, align=left] {Synthetic sequence};
	\draw [->] (Display) -- (Capture);
	
	\draw [->] (Capture) -- (Rectifiataion);
	\draw [->] (Rectifiataion) -- (bbox_calc);
	
	\draw [->] (bbox_calc) -- (pr_dataset);
\end{tikzpicture}
	}
	\caption{Fully automated pipeline for generating and annotating partly-real data.}
	\label{fig:pr_pipeline} 
\end{figure}
With the partly-real data the gap in terms of number, artifacts and degree of reality between synthetic and real image data can be closed. Even though the framework for generating synthetic data incorporates many effects and therefore bears many degrees of freedom as described in the previous section, it is still restricted compared to the real world. The partly-real data introduce additional artifacts as they occur for real data by acquiring the sequences with real-world cameras. Nevertheless, the recorded partly-real sequences can still be labeled fully automatic, which enables an easy and fast generation of large-scale data sets. \par 
The generation process of partly-real data is depicted in Figure~\ref{fig:pr_pipeline}. For the generation of partly-real sequences, rendered sequences, i.e. synthetic data as introduced in Section~\ref{sec:syn_data}, are used. These synthetic sequences are displayed on a large screen and recorded by real cameras. This procedure still allows controlled environmental conditions regarding illumination and perspective, but adds effects of real-world cameras to the sequences. These additional artifacts might be caused by cross-fading, camera lenses, sensor noise, automated white balance, auto exposure, or focus. Moreover, the cameras further post-process the data and store the recorded sequences in an encoded format. The coding of the sequences might introduce additional artifacts to the recorded data that are not yet present in the synthetic data set. %Moreover, additional coder configurations are added to the sequences by the cameras. 
\begin{figure}[t]
	\centering
	\includegraphics[width=0.8\columnwidth]{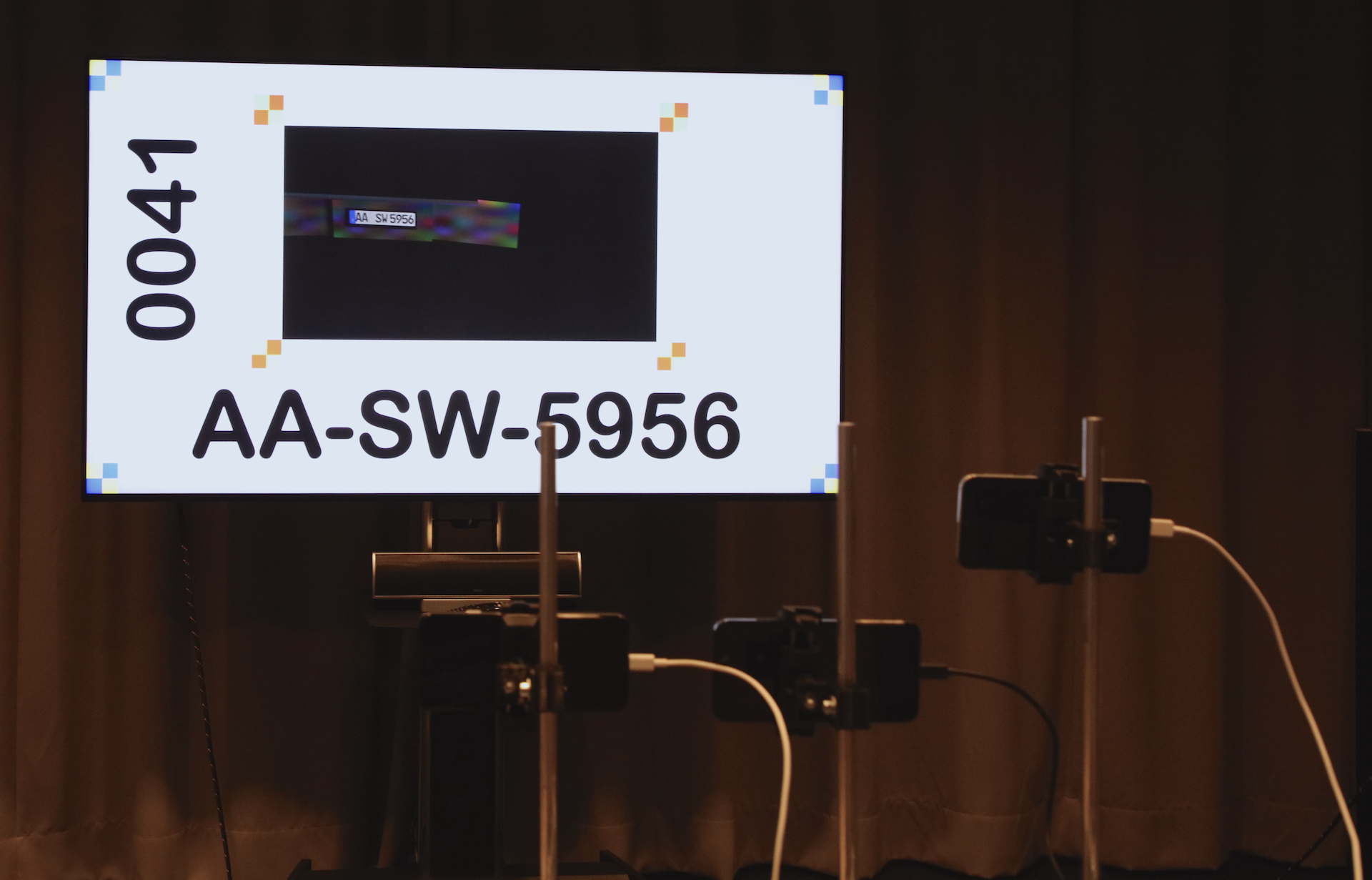}
	\caption{Recording setup of the partly-real data. The three mobile phones are mounted at a fixed position in front of the screen playing back the rendered sequences. The playback of the sequences and recording procedure of phones are controlled by computer. Additionally, the recorded sequences are transferred to the computer subsequently.}
	\label{fig:acq_pr_data}
\end{figure}
To generate sequences of different quality, three cameras are used for capturing the sequences. The cameras are equipped with different lenses, sensors, and pursue different post-processing schemes. As a low end camera, a {Nokia 2.2} smartphone is used. The mid range model is a {Samsung Galaxy A40}. As high end model, a {Google Pixel 4} is used. \par 
During the acquisition of the partly-real sequences, the cameras are mounted at a fixed position. All cameras record the same screen. The acquisition setup is depicted in Figure~\ref{fig:acq_pr_data}. Here, the synthetic sequences displayed on the screen have a resolution of $1920 \times 1080$~pixels. The synthetic sequences are embedded in a gray background with a resolution of $3840 \times 2160$~pixels. Left of the synthetic sequence, the currently displayed frame number is shown. Below of the synthetic sequence, the text label of the current license plate is denoted. After all synthetic sequences are prepared like that, the acquisition of the partly-real data is started. The entire acquisition process is automated. The playback of the first synthetic sequence is started by a central server as the recording is started on all three cameras. Once the playback of the sequence has ended, the recording is stopped as well. The recorded video file is subsequently transferred from each camera to the central server where it is stored. Hence, the cameras do not run out of storage space even for the acquisition of large data sets. During the entire acquisition procedure the cameras are connected to the server via USB and are controlled by the Android Debugging Bridge (ADB). During the recordings the room is kept dark, such that the screen showing the sequence is the only light source present. This guarantees stable illumination conditions throughout the recordings. Here, a $85"$ screen with 8K resolution is used for displaying the synthetic sequences during the acquisition of the partly-real data.\par
\begin{figure}[t]
	\centering
	\begin{subfigure}[]{0.48\columnwidth}
		\centering
		\includegraphics[width=\textwidth]{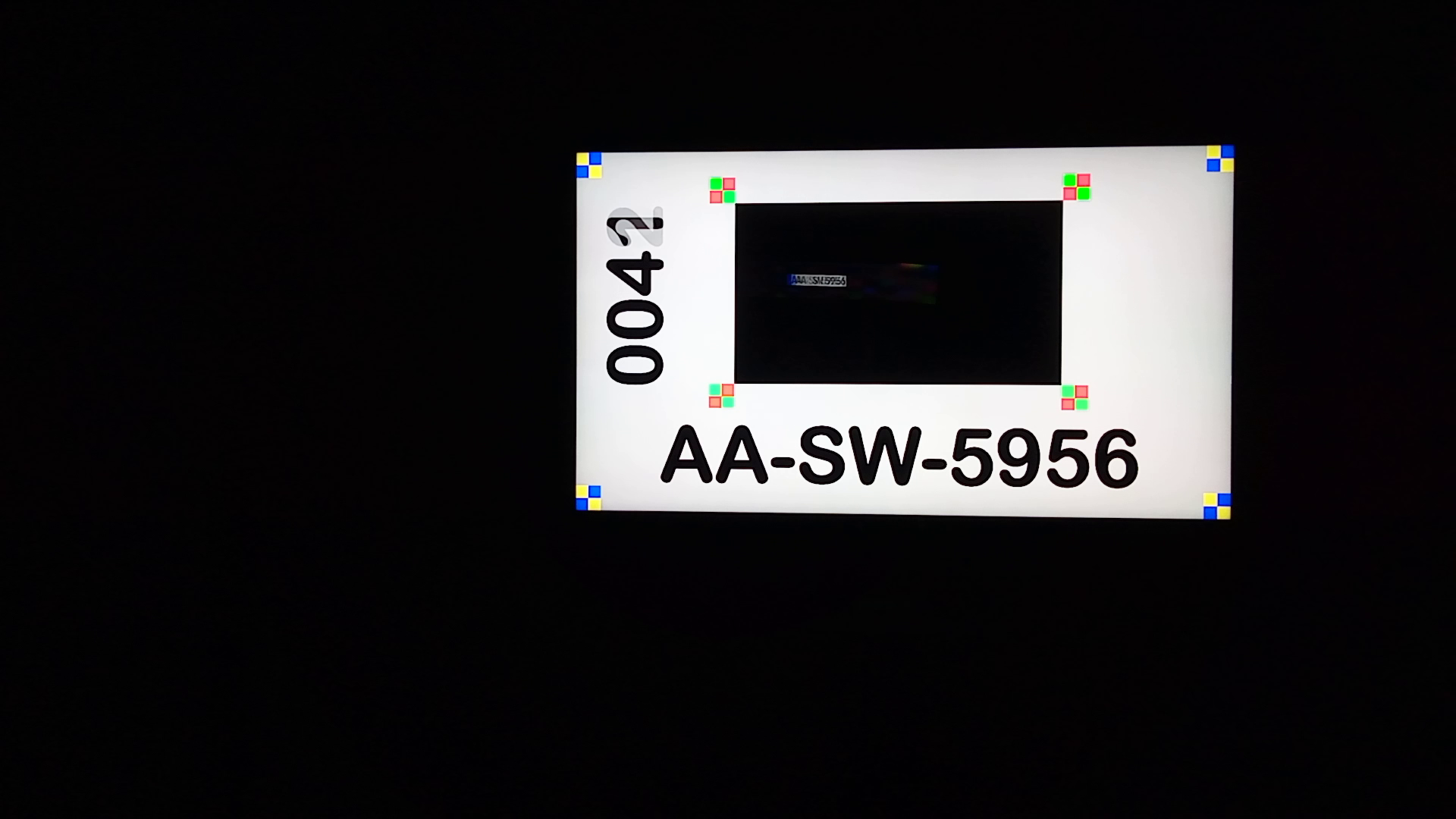}
		\caption{Nokia 2.2}
		\label{fig:nokia_raw} 
	\end{subfigure}
	\hfill
	\begin{subfigure}[]{0.48\columnwidth}
		\centering
		\includegraphics[width=\textwidth]{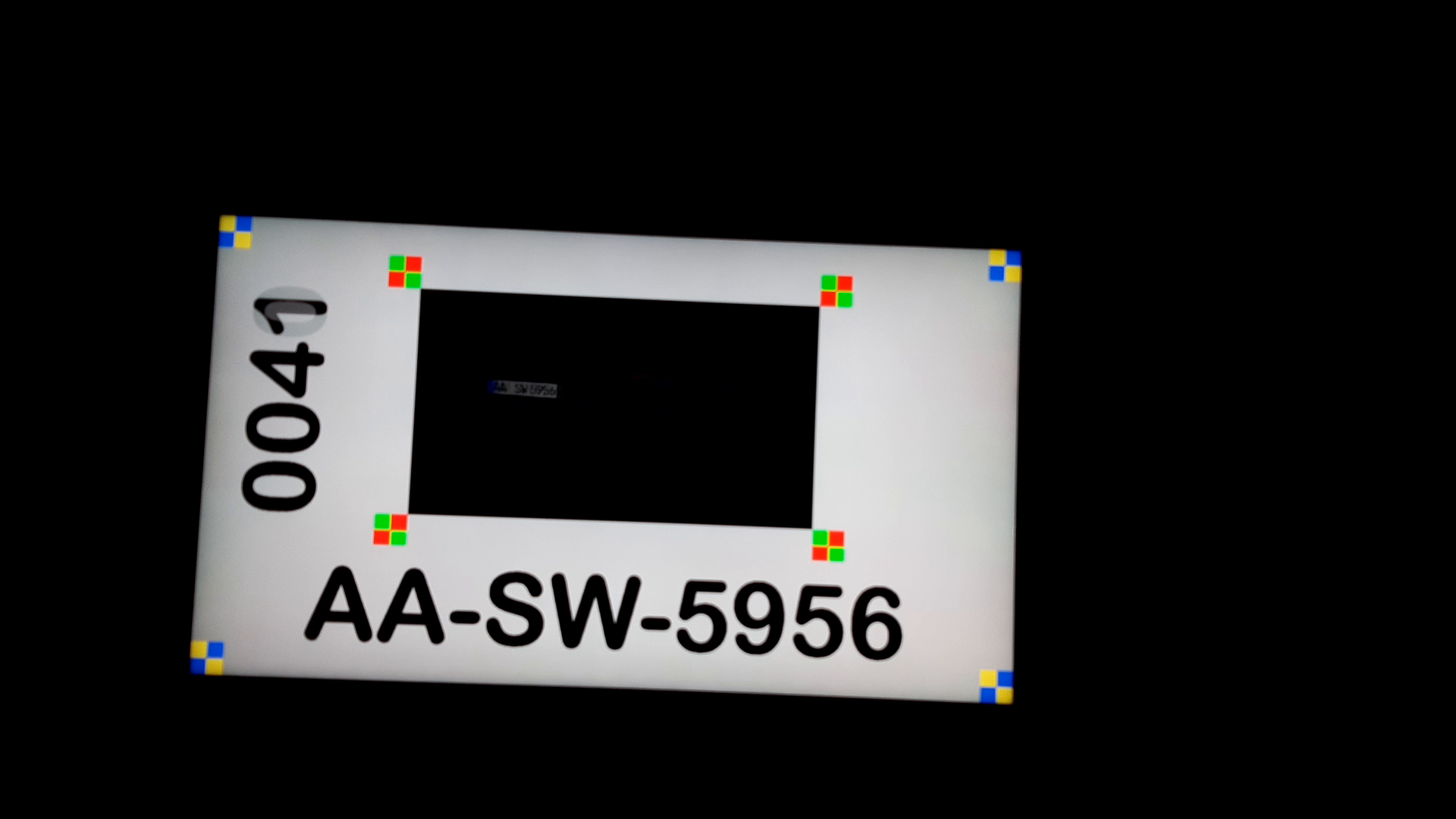}
		\caption{Samsung Galaxy A40}
		\label{fig:samsung_raw} 
	\end{subfigure}
	\begin{subfigure}[]{0.45\columnwidth}
		\centering
		\vspace{0.5cm}
		\includegraphics[width=\textwidth]{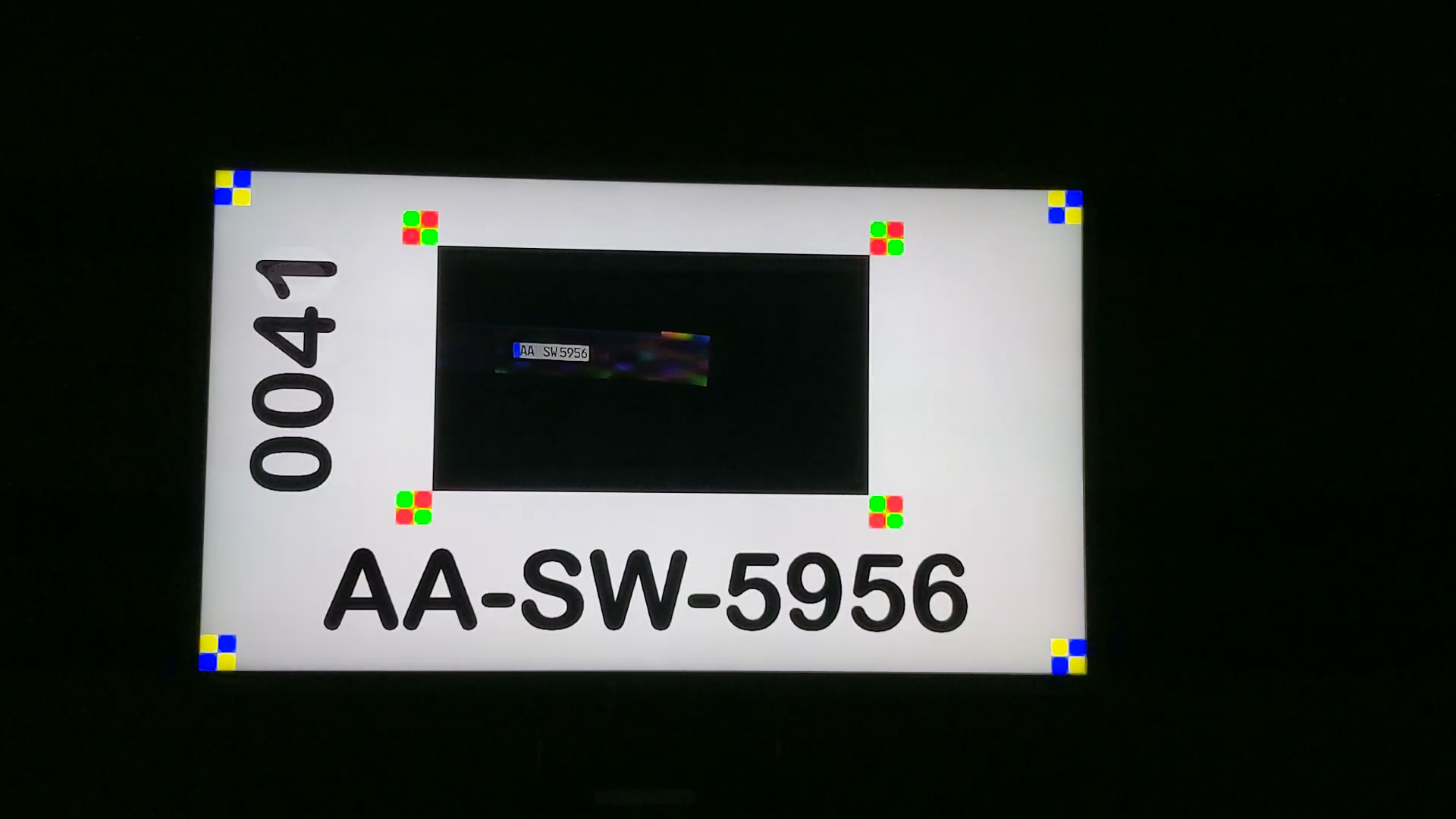}
		\caption{Google Pixel 4}
		\label{fig:pixel_raw} 
	\end{subfigure}
	\caption{Comparison of the same displayed frame captured with all three different phones. The raw data of the partly-real sequences is depicted.}
	\label{fig:example_all_phones}
\end{figure}

The three different cameras used capture different representation of the very same scene. Figure~\ref{fig:example_all_phones} depicts an example of the very same frame recorded by each of the cameras before post-processing. As can be seen, each camera produces a different output. The captured sequences differ in camera perspective and field of view. Additionally, the automatic white balance leads to different color representations among the recorded sequences. Also the frame rate of the recording influences the resulting image quality, as a lower recording frame rate leads to stronger cross-fading artifacts. \par
\begin{figure}[t]
\centering
\includegraphics[width=0.8\columnwidth]{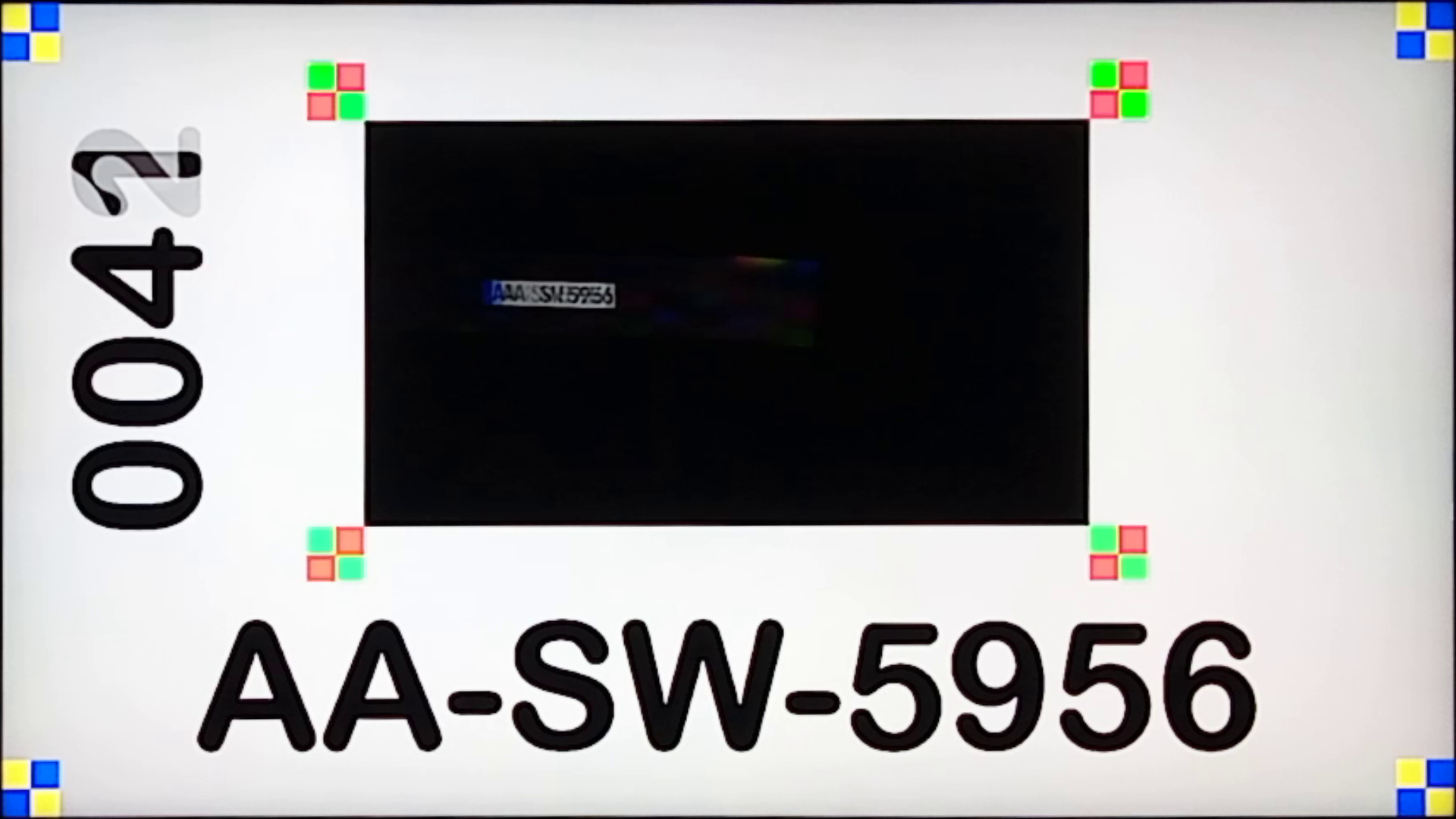}
\caption{Example of a rectified version of the displayed frame extracted from a partly-real image.}
\label{fig:rec_disp_frame}
\end{figure}

After the acquisition of the sequences, they are post-processed to annotate the data. This process is again fully automated. Therefore, the generation of labeled partly-real data is much faster than the generation of labeled real data. In a first step, the recorded video sequences are decoded in order to prepare them for the post-processing framework. During post-processing, the screen area is recognized within each frame. Once the area is detected, the homography is estimated. With the estimated homography the displayed frame is rectified and scaled to a size of $1920 \times 1080$~pixels. An example of the rectified screen area depicted in Figure~\ref{fig:nokia_raw} is shown in Figure~\ref{fig:rec_disp_frame}. \par
The position of the rendered sequence on the screen is always identical. Moreover, the position is known from the previous generation of the displayed frames. Therefore, the bounding box of the license plate can be easily calculated. Once the frame displayed on the screen is extracted and rectified, the calculation of the bounding box within the partly-real frame corresponds to a shift by a fixed offset and rescaling of the bounding box of the corresponding synthetic frame. In order to obtain the correct bounding box for the partly-real frames, the currently displayed frame number of the synthetic sequence has to be identified. This information is displayed on the left side of the screen. Using the Tesseract OCR engine \cite{Smith2007,Tesseract2019} the current frame number is recognized. If the post-processing framework fails at any step, i.e., extraction of the screen area or obtaining the current frame number, the current frame is skipped and the process continues with the next frame of the recorded sequence. \par 
Besides adding more degrees of freedom and additional artifacts to the sequences, the precision of the bounding box decreases when compared to the synthetic data. This is caused by inaccuracies during the estimations performed within the post-processing steps. Nevertheless, the annotations are still sufficiently precise and the variations can be regarded as implicit data augmentation. 
\subsection{Real Data Generation}
\label{sec:real_data}
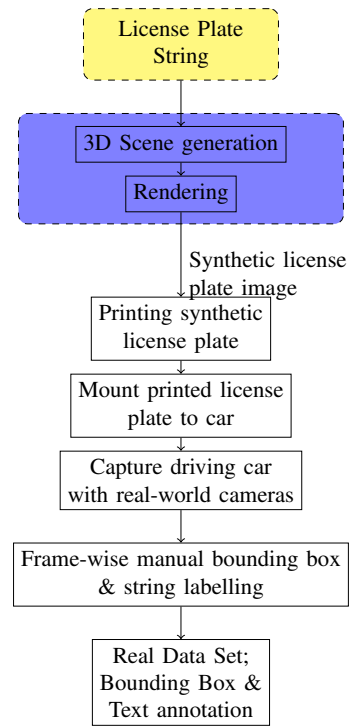
\begin{figure}[t]
	\centering
	\resizebox{0.9\columnwidth}{!}{
		\begin{tikzpicture}
	% Boxen
\node[minimum width=\columnwidth, minimum height=1.5cm] at (0,0) (Platzhalter) {}; % at (6,8) 
\node[draw,dashed,fill=yellow,fill opacity=0.5,rounded corners ,minimum width=3cm, minimum height=1.1cm] at (0,0) (csv) {}; % at (6,8) 

%	\node[align=right, anchor=north east, below =0cm of csv.south east] {csv-file};
	
	\node[align=center,minimum width=3cm, minimum height=1.1cm] at (0,0) (String) {License Plate\\String}; % at (3,8) 
%	\node[align=center,left= 0.2cm of String] (Scene_params) {Desired resolution,\\ desired scene length}; %  anchor=north west,
%	\node[align=center, right= 0.2cm of String] (Light) {Camera, light, and\\motion trajectory\\parameters};
	
	\node[draw,dashed,fill=blue,fill opacity=0.5,rounded corners ,minimum width=5cm, minimum height=1.7cm, below=0.5cm of String] (pipeline) {}; % at (6,8) 
%	\node[align=left, anchor=west, above right =0cm of pipeline.south east] {Rendering-based\\Pipeline};
	\node[draw, below=0.7cm of String] (Blender) {3D Scene generation}; %0.7cm 
	\node[draw, align = center, below=0.2cm of Blender] (Render) {Rendering};
	
	\node[draw, align = center, below=1.3cm of Render] (Print) {Printing synthetic\\license plate};
	\node[draw, align = center, below=0.2cm of Print] (Car) {Mount printed license\\plate to car};
	\node[draw, align = center, below=0.2cm of Car] (Capture) {Capture driving car\\with real-world cameras};
	
%	\node[draw, align = center, below=0.5cm of Capture] (Rectifiataion) {Extraction of screen area\&\\rectification};
	\node[draw, align = center, below=0.5cm of Capture] (Labeling) {Frame-wise manual bounding box\\ \& string labelling};
	
	\node[draw, align = center, below=0.5cm of Labeling] (real_dataset) {Real Data Set;\\Bounding Box \&\\Text annotation};
	
	% Pfeile
%	\draw [->] (Light) |- (Blender.east);
%	\draw [->] (Scene_params) |- (Blender.west);
	\draw [->] (String) -- (Blender);
	
	\draw [->] (Blender) -- (Render);
	
	\draw [->] (Render) -- (Print) node[near end, anchor=west, align=left] {Synthetic license\\plate image};
	\draw [->] (Print) -- (Car);
	\draw [->] (Car) -- (Capture);
	
	\draw [->] (Capture) -- (Labeling);
	
	\draw [->] (Labeling) -- (real_dataset);
\end{tikzpicture}
	}
	\caption{Pipeline used for acquiring real data. The generation of samples is again automatized, the acquisition and annotation of the sequences has to be done manually.}
	\label{fig:real_pipeline} 
\end{figure}
The proposed rendering-based data generation pipeline can as well aid the acquisition of real data. The generation of this type of data is the most elaborate of all three presented data types. The generation procedure is depicted in further detail for license plate data in Figure~\ref{fig:real_pipeline}. Here, no previous knowledge about the location of the license plates within the sequence can be exploited as the real-world recording conditions are unconstrained. For the acquisition of the data, different samples are required. For the acquisition of license plates, different samples are rendered that hold randomly generated text configurations as for the synthetic and partly-real data. Again, the random text configurations ensure conformity with current regulations regarding data privacy. \par
The samples are rendered from a frontal perspective eliminating all geometric distortions. The illumination of the 3D scene is constant which avoids shadowing. Instead of a whole sequence only a single frame is rendered. Therefore, the parameters specifying the scene length, camera, light and motion trajectory can be omitted during the generation of the configuration file. In a subsequent post-processing step, the rendered frame is cropped such that it only shows the object without any background from the scene using the bounding box annotation of the synthetic frame. \par 
For license plates, the cropped images are printed matching the size of real-world German license plates, i.e., 52 cm in width and 11 cm in height. The printed versions of the license plates are mounted in the front and back of cars at the location where it is specified by the manufacturer. During the acquisition of the real sequences the prepared cars drive past a stationary setup holding several different cameras.  \par
\begin{figure}[t]
\centering
\includegraphics[width=0.8\columnwidth]{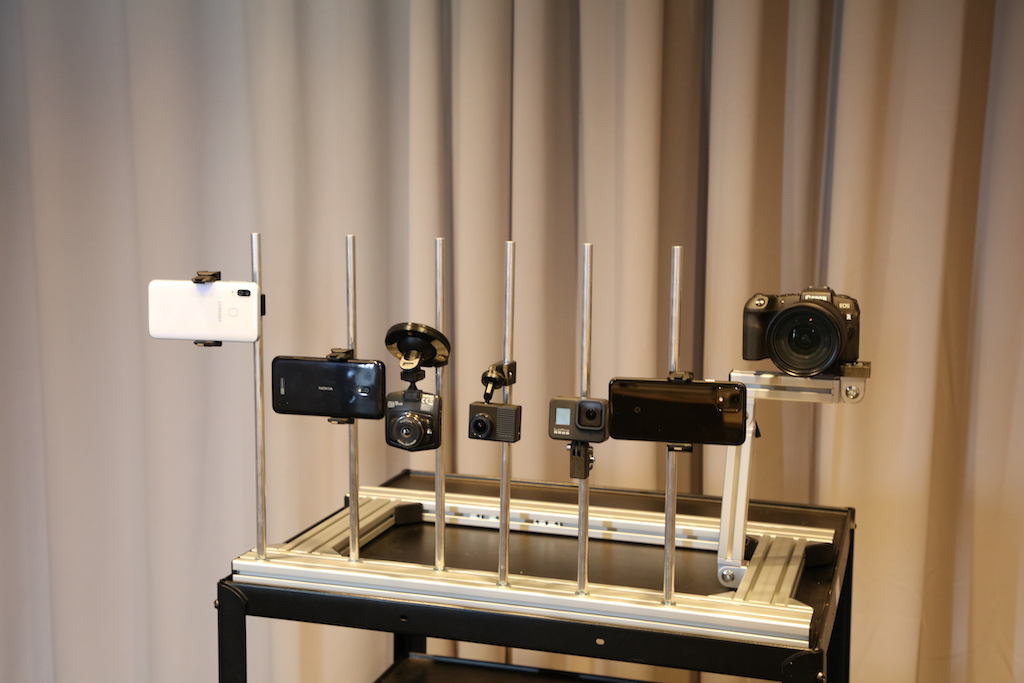}
\caption{Setup for recording the real data.}
\label{fig:camera_setup}
\end{figure}

 %
An image of the camera setup used for the acquisition can be seen in Figure~\ref{fig:camera_setup}. It holds in total seven different cameras. Among these cameras are the three cameras that were already used for capturing the partly-real data (Nokia 2.2, Samsung A40, Google Pixel~4) representing different price ranges. Furthermore, two dashcams (Denver CCT-1210, Garmin DC~66W) from a low and medium price range are used. Moreover, a GoPro Hero~8 action camera as well as a high quality Canon EOS~RP equipped with a Canon RF 24-105mm F4 L lens are used. The recordings are conducted on closed private property. Therefore, only samples produced by the proposed rendering-based data generation pipeline are acquired. \par 
The camera setup is positioned such that it captures the moving samples, i.e. the license plates mounted to the driving cars, as long as possible. During the recording, it is taken care that only one sample is present within each sequence, i.e. there are no further cars and license plates in the background. The recorded sequences are encoded and stored by each camera according to its individual processing scheme. After the recording, the sequences are cut such that the current sample is visible from the beginning on. The sequences are cut at Intra-frames only. This prevents a re-encoding of the sequences. After preparing every sequence like that, the resulting sequences are labeled manually, i.e. each frame of the sequences is annotated. For the application of license plate recognition, the annotation holds the text string printed on the current license plate sample. This text label is the same throughout each sequences. Therefore, the text label does not change within a sequence. Besides the text string, also the position of the license plate sample within the current frame is annotated. Therefore, a minimal enclosing rectangle is drawn manually around the license plate in each frame. Moreover, it is denoted if the license plate sample is occluded. Occlusions might occur by the car entering or leaving the current field of view of the camera. \par 
Figure~\ref{fig:real_LP} shows two cropped images of the same license plate sample from the real data set. Here it can be seen that the quality of the same license plate sample may change drastically over the duration of a sequence, as the distance of the sample, the viewing angle, and the illumination may change. While the license plate is still readable in the upper line of Figure~\ref{fig:real_LP}, the visual quality is severely impaired for the sample in the bottom line of Figure~\ref{fig:real_LP}. By incorporating such samples into the generated data set, challenging application scenarios can be addressed as intended during the experimental validation of the proposed data generation pipeline. 
\section{Experimental Setup}
\label{sec:exp_setup}
During the experimental validation of the proposed rendering-based data generation pipeline, we will examine the application of automatic license plate recognition. Here, we focus on the recognition of samples with impaired visual quality, as this is still an open research topic. Recognition of visually impaired samples is of high importance in forensic applications. \par
The detection of license plates is nowadays commonly performed by NNs~\cite{Silva2018,Hsu2017,Chan2020}. For the training of these detectors, license plate data sets can be used that do not necessarily show license plates from the desired country, as the dimensions of the license plates are commonly similar. Therefore, the experiments conducted here focus on the recognition of license plates, as this is bound to samples from the desired region. Here, we will assess the recognition performance of a NN-based OCR application, trained on data synthesized with our proposed rendering-based pipeline. During the evaluation we also examine the influences of incorporating synthetic and partly-real data during the training of the OCR algorithm. Additionally, we examine the size of the incorporated data sets of different data types and their influence onto the recognition performance. Throughout the experiments carried out here, the recognition of license plates with impaired visual quality is of special interest. \par 
For the recognition of the license plates, the commonly used open source OCR algorithm Tesseract~\cite{Smith2007} was used. Here, Tesseract v4.1.1 is used during the experiments. This Tesseract version uses an LSTM for the recognition of the characters within the text line~\cite{Tesseract2019}. The Tesseract OCR engine is not only used for the recognition of natural text but also used for the recognition of text on license plates. The open source license plate recognition software OpenALPR \cite{Hill2016} for example uses Tesseract as application for the character recognition. In~\cite{Hegghammer2021} it was shown that Tesseract can compete with current commercial, server-based OCR algorithms such as Amazon Textract and Google DocumentAI for noise-free images of English text. The usage of server-based OCR algorithms is often prohibited for ALPR applications due to restrictions regarding data protection and privacy, as the confidentially cannot be guaranteed for cloud-based applications~\cite{Hegghammer2021}. Moreover, server-based applications cannot be retrained by the user to match the individual use case. The data synthesized by the proposed rendering-based pipeline can as well be used for any other application, as the data set is fully labeled. \par 
As the proposed pipeline for synthesizing data is capable of producing three different types of data, the effect of the individual types of data are evaluated separately. Each type of data is subdivided individually into a training, validation, and test split. Moreover, the training splits of the synthetic and partly-real data are additionally subdivided into subsets holding 100, 75, 50, and 25 percent of the available training data of the respective data type. Furthermore, additional synthetic and partly-real subsets are built that have the same size as the real data set holding $0.6\%$ and $0.9\%$ of the full synthetic and partly-real training sets respectively. These sets are denoted as subset 0 here. The exact size of the different data sets is given in Table~\ref{tab:data_distribution}. The largest compartment of the database synthesized by the proposed rendering-based pipeline is the synthetic data set. The full set holds over 1.5 million training and over 175,000 validation samples. Here, all synthetic samples hold unique text labels. The second part of the database generated by the proposed rendering-based pipeline holds partly-real data. As introduced in Section~\ref{sec:pr_data}, the partly-real data bring the synthetic sequences closer to reality. This is achieved by incorporating real acquisition systems, while bearing the advantages of synthetic data regarding the automated labeling process. As the acquisition procedure of the partly-real data set is more elaborate than generating synthetic data, the number of partly-real samples is smaller than for the synthetic samples. The full partly-real data set holds over 950,000 training and over 100,000 validation samples. The full partly-real data set holds license plates with 2000 unique text labels. The third part of the database holds samples of license plates acquired under real-world conditions. As the real data set has to be labeled manually, it builds the smallest of the three data sets. The real data set holds 9,040 training and 1,005 validation samples. Furthermore, 647 samples with a height of at least 10 pixels are available for testing. The samples originate from license plates with 100 unique text labels. The synthetic and partly-real subsets are generated by randomly excluding samples from the full training set, i.e., subset 100. The test sets contain only samples and text strings that are not used within any of the other data sets. The test sets remain unchanged throughout all conducted experiments. \par
\begin{figure*}[t]
	\centering
	\begin{subfigure}[]{0.3\textwidth}
		\centering
		\includegraphics[width=\textwidth]{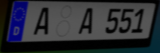}
		\caption{Synthetic.}
		\label{fig:syn_rot_comp}
	\end{subfigure}
	\hfill
	\begin{subfigure}[]{0.3\textwidth}
		\centering
		\includegraphics[width=\textwidth]{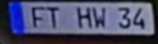}
		\caption{Partly-real.}
		\label{fig:pr_rot_comp}
	\end{subfigure}
	\hfill
	\begin{subfigure}[]{0.3\textwidth}
		\centering
			\includegraphics[width=\textwidth]{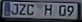} \\
			\vspace{0.2cm}
			\includegraphics[width=\textwidth]{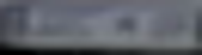}
		\caption{Real data.}
		\label{fig:real_LP}
	\end{subfigure}
	\caption{Comparison of the different samples produced by the proposed rendering-based data generation pipeline. The depicted samples are prepared for training the OCR application by cropping the generated images and applying a rotation compensation for the synthetic and partly-real data. The samples from the real data set show the very same sample acquired from different distances. With growing distance to the camera the visual quality of the sample degrades.}
	\label{fig:comp_prep_samples}
\end{figure*}

 
\begin{table}[t]
	\centering
	\caption{Data set distribution giving the number of samples per split.} 
	\begin{tabular}{|l|c|c|c|c|c|}
		\hline
		& Subset & Training & Validation & Test & Total \\ \hline
		\multirow{5}{*}{Synthetic} & 0 & 9040 & 1.004 & \multirow{5}{*}{189.158} & 199.202 \\ \cline{2-4} \cline{6-6}
		& 25 & 394.007 & 43.779 & & 626.944 \\ \cline{2-4} \cline{6-6}
		& 50 & 788.014 & 87.558 & & 1.064.730 \\ \cline{2-4} \cline{6-6}
		& 75 & 1.182.022 & 131.336 & & 1.502.516 \\ \cline{2-4} \cline{6-6}
		& 100 & 1.576.029 & 175.115 & & 1.940.302 \\ \hline
		\multirow{5}{*}{Partly-real} & 0 & 9.040 & 1.004 & \multirow{5}{*}{118.238} & 128.282 \\ \cline{2-4} \cline{6-6}
		& 25 & 238.732 & 26.526 & & 383.496 \\ \cline{2-4} \cline{6-6}
		& 50 & 477.464 & 53.052 & & 648.754 \\ \cline{2-4} \cline{6-6}
		& 75 & 716.195 & 79.578 & & 914.011 \\ \cline{2-4} \cline{6-6}
		& 100 & 954.927 & 106.104 & & 1.179.269 \\ \hline
		{Real} & --- & 9.040 & 1.005 & 647 & 10.692\\ \hline
	\end{tabular}
	\label{tab:data_distribution}
\end{table}
For the training of the OCR application, the samples are preprocessed by a rotation compensation and resizing to a uniform height. For the synthetic data, an exact bounding box of the license plate is known from the rendering process. Using this bounding box, the images are cropped such that they only hold the license plate. Therefore, a rotated minimal enclosing rectangle around the bounding box is estimated. With the rotated enclosing rectangle, a rotation matrix is estimated. Using this matrix, the images are warped coping for the rotational aspects. Within this procedure only the rotational aspects are removed, the remaining geometric transformation aspects are still present within the prepared images. An example of an image prepared following this procedure is depicted in Figure~\ref{fig:syn_rot_comp}. Even though, the rotational transformation is compensated, the samples may still contain e.g. a shearing transformation as depicted in Figure~\ref{fig:syn_rot_comp}. \par 
For the generation of the partly-real data, synthetic sequences with a duration of 50 frames were rendered. The sequences contain linear motion of the license plate through the current field of view. These synthetic sequences were acquired and post-processed as described in Section~\ref{sec:pr_data}. As for the synthetic data set, the rotation of the single samples from the partly-real data set is compensated. This again only removes the rotational aspects while remaining all other geometric transformation aspects. An example of an image prepared following this procedure is depicted in Figure~\ref{fig:pr_rot_comp}. \par 
Using the proposed rendering-based data generation pipeline also real data is acquired following the procedure described in Section~\ref{sec:real_data}. This data set is labeled entirely by hand. As the sequences contain cars that drive past the camera system, the size and visual quality of the license plate change over the duration of the sequence. Thereby, also the viewing angle under which the samples are recorded changes over time. Therefore, the recognition of many samples from the acquired data set is very challenging. This challenging recognition scenario that also includes many samples of visual impaired quality represents a forensic license plate recognition application well. Forensic applications differ from common civil applications as the considered samples are commonly acquired under adverse conditions. Therefore, common applications often fail to perform well for forensic scenarios and have to be retrained for the desired task. \par
Compared to the synthetic and partly-real data set, the real data set seems rather small. However, when compared to other state-of-the-art databases as given in Table~\ref{tab:sota_datasets_sizes} it can be observed that its size is similar or even larger than other common data sets. For example the size of the commonly used AOLPE or PKU Vehicle data set is less than half of our real data set. The ground-truth annotation is identically structured for all three types of data. It does not contain any blanks. The region identifier is separated from the individual identifier by a dash representing the larger margin between the first and second text block. \par 
The recognition results are evaluated in terms of character error rate (CER). The CER measures the number of wrongly recognized characters in relation to the overall number of characters $N_C$ within the current string. The number of falsely recognized characters is obtained by the Levenshtein distance $d_{\mathrm{Lev}}\left(\hat{L},L\right)$. The Levenshtein distance gives the minimal number of editing steps that are required to transform the recognized text string $\hat{L}$ into the ground-truth text string $L$. This makes the Levenshtein distance robust against insertions, substitutions and deletions. Thereby, the CER is defined as 
\begin{equation}
\mathrm{CER} = \frac{d_{\mathrm{Lev}}\left(\hat{L},L\right)}{N_\mathrm{C}}.
\end{equation}
Moreover, the miss rate (MR) is evaluated. The MR denotes the ratio of wrongly recognized license plates. Here, a license plate is regarded as wrongly recognized whenever at least one character is not correctly recognized. The MR is defined as 
\begin{equation}
	\mathrm{MR} = \frac{N_\mathrm{F}}{N_\mathrm{T}+N_\mathrm{F}},
\end{equation}
where $N_\mathrm{T}$ denotes the number of samples recognized entirely correct and $N_\mathrm{F}$ denotes the number of wrongly recognized samples.
\section{Evaluation}
\label{sec:eval}
For the evaluation of the proposed framework, the different types of data are used for training the Tesseract OCR engine. For the experiments conducted here, especially challenging scenarios regarding viewing angle, size of the license plate or illumination were examined. These scenarios are of special interest with regard to forensic applications. In forensic applications the considered samples often show an impaired visual quality, as the camera is commonly not directed towards the license plate and the desired sample is only visible in the background. This often leads to a challenging viewing angle, bad resolution or insufficient illumination. Moreover, the license plates are often blurred or corrupted by other artifacts. These challenging characteristics compromise the recognition performance of ALPR systems. 
\subsection{Reference}
The common approach to alleviate this problem is to train the OCR application from scratch by using a specific data set of license plates from the region of interest that includes similar artifacts. Due to restrictions regarding data protection and the elaborate labeling process, data sets holding real world training data are very limited. The size of the real data set produced with our proposed rendering-based data generation is comparable to common real-world license plate data sets, as denoted in Table~\ref{tab:sota_datasets_sizes}. Using our generated real data set, we trained the Tesseract OCR application in its version 4.1.1 for three epochs. Thereby, a CER of $73.74\%$ and a MR of $100\%$ is obtained on the independent real-world test set. The very high CER and MR indicate that the recognition scenario is too challenging for an algorithm trained directly on the real data set. Moreover, this shows that the scenario of forensic license plate recognition is more challenging than license plate recognition for civil applications such as access control where the considered samples hold notably less artifacts. The results suggests that the real data set is  too small to train the algorithm on it solely, as none of the license plates is recognized correctly and the large majority of characters is recognized falsely, as well.\par 
\begin{table*}[t]
	\centering
	\caption{CER and MR for the algorithm trained for three epochs on the individual training data splits and evaluated on the real test data set.} 
	\begin{tabular}{|l|cc|cc|cc|cc|cc|}
		\hline
		{Training Set} &  \multicolumn{2}{c|}{0} &  \multicolumn{2}{c|}{25} &   \multicolumn{2}{c|}{50} &   \multicolumn{2}{c|}{75} &   \multicolumn{2}{c|}{100}\\ \cline{2-11}
		 & CER [\%] & MR [\%] & CER [\%] & MR [\%] & CER [\%] & MR [\%] & CER [\%] & MR [\%] & CER [\%] & MR [\%] \\ \hline
		 Real & -- & -- & -- & -- & -- & -- & -- & -- & 73.74 & 100 \\ \hline% \cline{2-11}
		 Synthetic & 55.78 & 89.03 & 48.36 & 85.16 &  40.30 & 74.50 &  \textbf{39.31} & \textbf{73.11} &  45.31 & 82.53  \\ \hline% \cline{2-11}
		 Partly-real   &  34.83 & 80.22  &  21.28 & 55.49  &  \textbf{19.04} & 50.70  & 20.04 & 53.79  &  21.76 & \textbf{47.60} \\ \hline% \cline{2-11}
		Synthetic + Partly-real &  31.27 & 68.16  &  18.89 & 46.21  &  17.52 & 48.69  &  \textbf{14.11} & \textbf{41.27} &  19.43 & 47.30  \\ \hline
	\end{tabular}
	\label{tab:results_syn_pr_syn+pr}
\end{table*}
\subsection{Rendering-based Pipeline}
Therefore, training the OCR application on the synthetic and partly-real data sets, synthesized by the proposed data generation pipeline, is further examined. By evaluating the obtained CER and MR on the independent real-world test set, it can be observed that training the OCR application on the synthetic and partly-real data set can decrease the error rates significantly. Table~\ref{tab:results_syn_pr_syn+pr} denotes the results for training Tesseract on each of the three data types synthesized by the proposed pipeline for three epochs. Moreover, here the evaluation results for training Tesseract on the subsets of different size are listed. The data type incorporated during training is specified line-wise in Table~\ref{tab:results_syn_pr_syn+pr}. Here, synthetic indicates that only synthetic data is used during training. Analogously, partly-real denotes that only partly-real data is incorporated during training. The last line in Table~\ref{tab:results_syn_pr_syn+pr} denotes the results obtained with a Tesseract application trained on the combination of the respective synthetic and partly-real data set. Each trained Tesseract version is evaluated on the real-world test data set as this type of data is used in the final application as well. The performance of the algorithms trained on the differently sized training subsets is denoted column-wise in Table~\ref{tab:results_syn_pr_syn+pr}. As the real-world training data set is not further subdivided, the first line holds only entries for the full real-world training set. As already discussed, the obtained error rates for training the algorithm on the real training data set are too high to use this algorithm effectively in an ALPR application. Therefore, the OCR algorithm is trained on synthetic data generated by the proposed rendering-based data generation pipeline. As shown in the table, the best obtained CER for Tesseract trained on synthetic data solely is $39.31\%$, the lowest MR is $73.11\%$. This already denotes a drastic reduction of the error rates, compared to training on real-world data only. %By training on synthetic data only, the CER can be reduced by up to 34.43 percentage points and the MR can be reduced by 26.89 percentage points in the best case of using subset 75. 
By further investigating Table~\ref{tab:results_syn_pr_syn+pr}, it can be observed that training the OCR application on a synthetic data set of the same size as the real data set is already beneficial. Here the error rates could as well be reduced notably to $55.78\%$ and $89.03\%$ in terms of CER and MR respectively. \par
When training the algorithm on the partly-real data, the error rates can be decreased even further. By using the partly-real data, the CER can be reduced to $19.04\%$ and the MR to $47.60\%$ for the partly-real training set 50 and 100, respectively. \par% This corresponds to a reduction of $54.70$ percentage points in terms of CER and $52.54$ percentage points in terms of MR compared to a training on real data only. \par 
By combining the synthetic and partly-real data sets, the error rates can be decreased even further. By the combination of both data types, a CER of $14.11\%$ and a MR of $41.27\%$ can be achieved for the training set 75. %These error rates mark a reduction of $59.63$ and $58.73$ percentage points in terms of CER and MR, respectively. 
It is important to note that the results denoted in Table~\ref{tab:results_syn_pr_syn+pr} for the synthetic, partly-real, and the combination of both data types are obtained without incorporating any real data during training. Thereby, our proposed rendering-based data generation pipeline offers an option to train NNs even for application scenarios where no real data is available. The entire generation and annotation process of the synthetic and partly-real data is automated. Due to this an elaborate acquisition and labeling procedure is obsolete for training a well performing NN.\par 
\begin{table*}[t]
	\centering
	\caption{CER and MR for the algorithm trained for three epochs on the individual training data splits, subsequently finetuned for one epoch on the real-world data set and evaluated on the real test data set.} 
	\begin{tabular}{|l|cc|cc|cc|cc|cc|}
		\hline
		{Training Set} &  \multicolumn{2}{c|}{0} &  \multicolumn{2}{c|}{25} &   \multicolumn{2}{c|}{50} &   \multicolumn{2}{c|}{75} &   \multicolumn{2}{c|}{100}\\ \cline{2-11}
		& CER [\%] & MR [\%] & CER [\%] & MR [\%] & CER [\%] & MR [\%] & CER [\%] & MR [\%] & CER [\%] & MR [\%] \\ \hline
		Real & -- & -- & -- & -- & -- & -- & -- & -- & 73.74 & 100 \\ \hline% \cline{2-11}
		Synthetic & 57.56 & 100.00 & \textbf{22.55} & 69.55 & 25.96 & 67.39 & 24.51 & \textbf{62.44} &  29.37 & 70.48 \\ \hline% \cline{2-11}
		Partly-real   & 53.80 & 100.00 & 15.99 & 56.11 & 14.26 & \textbf{41.42}  & 14.87 & 47.14 & \textbf{11.90} & 41.89 \\ \hline% \cline{2-11}
		Synthetic + Partly-real & 40.76 & 94.59 & 14.37 & 42.19 & 13.02 & 41.58 & 14.42 & 50.23 & \textbf{12.71} & \textbf{39.88}  \\ \hline
	\end{tabular}
	\label{tab:results_syn_pr_syn+pr_ft_on_real}
\end{table*}
\subsection{Finetuning}
The previously shown results demonstrate the usability of the proposed rendering-based data generation pipeline for scenarios where no real training data is available. However, for scenarios as the one described here, where at least a limited amount of real training data is available, the results shown before and given in Table~\ref{tab:results_syn_pr_syn+pr} can be further improved. Therefore, the OCR applications trained for the previous experiments on the synthetic, partly-real, and the combined training sets, are finetuned using the real training data set. The results obtained for the finetuned models evaluated on the real-world test set, are denoted in Table~\ref{tab:results_syn_pr_syn+pr_ft_on_real}. The table is structured analog to Table~\ref{tab:results_syn_pr_syn+pr}. The differently sized training subsets are given column-wise. The data type used for the initial training is denoted line-wise. After the training on the artificially generated synthetic and partly-real data, each of the trained versions is finetuned for one epoch on the real training data set. \par 
The first line of Table~\ref{tab:results_syn_pr_syn+pr_ft_on_real} again holds the baseline reference of Tesseract trained only on the real training data set solely. The second line denotes the results for the algorithm trained on the different synthetic data sets in a first step and subsequently finetuned on the real training data set. The second line is produced analogously for the partly-real data. The results denoted in the last line are obtained on a Tesseract version trained on the combined synthetic and partly-real training data sets and subsequently finetuned on the real training data set.  By the additional finetuning, the error rates can be decreased further. For the algorithm trained on the synthetic data the CER can be decreased to $22.55\%$ and the MR to $62.44\%$ after finetuning the model trained on training set 25 and 75 respectively. For the algorithm trained on the partly-real data the CER is reduced to $11.90\%$ and the MR to $41.42\%$ for training set 100 and 50 respectively by a subsequent finetuning. For the algorithm trained initially on the combined training subset 100 of synthetic and partly-real data, the CER is reduced to $12.71\%$ and the MR to $39.88\%$. \par %Thereby, the error rates can be reduced by $61.84$ percentage points in terms of CER compared to the baseline model by training on the partly-real training set 100 and finetuning on the real training data set. The MR can be decreased by $60.12$ percentage points by finetuning the the algorithm trained on the subset 100 holding the combination of synthetic and partly-real data, compared to Tesseract trained on the real data solely.\par 
\begin{figure*}[t]
	\centering
	\begin{subfigure}[]{0.9\columnwidth}
		\centering
		\resizebox{\textwidth}{!}{
\definecolor{mycolor1}{rgb}{0.00000,0.44700,0.74100}%
\definecolor{mycolor2}{rgb}{0.85000,0.32500,0.09800}%
\definecolor{mycolor3}{rgb}{0.92900,0.69400,0.12500}%
\definecolor{mycolor4}{rgb}{0.49400,0.18400,0.55600}%
\begin{tikzpicture}

\begin{axis}[%
width=4.521in,
height=3in,
scale only axis,
xmin=0,
xmax=100,
xtick distance = 25, % manually added
xlabel style={font=\color{black}},
xlabel={{Subset}},
ymin=0,
ymax=105,
ytick distance = 10, % manually added
ylabel style={font=\color{black}},
ylabel={{CER in \%}},
%axis background/.style={fill=white},
legend columns=3, 
legend style={legend cell align=left, align=left, align=left, at={(1,1)}, anchor=north west, align=left, at={(0.5,-0.25)}, anchor=center, draw=black},
mark size=1mm
]
\addplot [color=mycolor1, mark=*, line width = 0.7mm]
  table[row sep=crcr]{%
100 45.31\\
75 39.31\\
50 40.30\\
25 48.36\\
0 55.78\\
};
\addlegendentry{Synthetic}

\addplot [color=mycolor1,dotted, mark=*, line width = 0.7mm]
table[row sep=crcr]{%
	100													29.37\\
	75													24.51\\
	50													25.96\\
	25													22.55\\
	0													57.56\\
};
\addlegendentry{Synthetic, FT}

%\addplot [color=mycolor1,dashed, mark=*, line width = 0.7mm]
%table[row sep=crcr]{%
%	100																				29.82\\
%	75																				23.95\\
%	50																				25.45\\
%	25																				28.76\\
%	0																				72.16\\
%};
%\addlegendentry{Synthetic, FT 3Ep}

\addplot [color=mycolor2, mark=triangle, line width = 0.7mm]
  table[row sep=crcr]{%
100						21.76\\
75						20.04\\
50						19.04\\
25						21.28\\
0						34.83\\
};
\addlegendentry{Partly-real}

\addplot [color=mycolor2,dotted, mark=triangle, line width = 0.7mm]
table[row sep=crcr]{%
	100													11.90\\
	75													14.87\\
	50													14.26\\
	25													15.99\\
	0													53.80\\
};
\addlegendentry{Partly-real, FT}

%\addplot [color=mycolor2,dashed, mark=triangle, line width = 0.7mm]
%table[row sep=crcr]{%
%	100																				14.39\\
%	75																				19.12\\
%	50																				17.54\\
%	25																				23.32\\
%	0																				62.93\\
%};
%\addlegendentry{PR, FT 3Ep}

%\addplot [color=mycolor3, mark=*, dash pattern=on 1pt off 3pt on 3pt off 3pt, line width = 0.7mm]
\addplot [color=mycolor3, mark=diamond, line width = 0.7mm]
  table[row sep=crcr]{%
100						19.43\\
75						14.11\\
50						17.52\\
25						18.89\\
0						31.27\\
};
\addlegendentry{Synthetic+Partly-real}

\addplot [color=mycolor3, dotted, mark=diamond, line width = 0.7mm]
table[row sep=crcr]{%
	100													12.71\\
	75													14.42\\
	50													13.02\\
	25													14.37\\
	0													40.76\\
};
\addlegendentry{Synthetic+Partly-real, FT}

%\addplot [color=mycolor3, dashed, mark=diamond, line width = 0.7mm]
%table[row sep=crcr]{%
%	100																				13.27\\
%	75																				18.26\\
%	50																				19.25\\
%	25																				16.35\\
%	0																				57.39\\
%};
%\addlegendentry{Syn+PR, FT 3Ep}

%\addplot [color=mycolor4, mark=*, dash pattern=on 3pt off 6pt on 6pt off 6pt, line width = 0.7mm]
\addplot [color=mycolor4, line width = 0.7mm]
  table[row sep=crcr]{%
100	73.74 \\
75	73.74 \\
50	73.74 \\
25	73.74 \\
0	73.74 \\
};
\addlegendentry{Baseline (real only)}

\end{axis}
\end{tikzpicture}%
}
	\caption{CER}
	\label{fig:CER}
\end{subfigure}
%\hfill
\qquad
\begin{subfigure}[]{0.9\columnwidth}
	\centering
	\resizebox{\textwidth}{!}{
\definecolor{mycolor1}{rgb}{0.00000,0.44700,0.74100}%
\definecolor{mycolor2}{rgb}{0.85000,0.32500,0.09800}%
\definecolor{mycolor3}{rgb}{0.92900,0.69400,0.12500}%
\definecolor{mycolor4}{rgb}{0.49400,0.18400,0.55600}%
\begin{tikzpicture}

\begin{axis}[%
width=4.521in,
height=3in,
scale only axis,
xmin=0,
xmax=100,
xtick distance = 25, % manually added
xlabel style={font=\color{black}},
xlabel={{Subset}},
ymin=0,
ymax=105,
ytick distance = 10, % manually added
ylabel style={font=\color{black}},
ylabel={{MR in \%}},
%axis background/.style={fill=white},
legend columns=3, 
legend style={legend cell align=left, align=left, align=left, at={(1,1)}, anchor=north west, align=left, at={(0.5,-0.25)}, anchor=center, draw=black},
mark size=1mm
]
\addplot [color=mycolor1, mark=*, line width = 0.7mm]
  table[row sep=crcr]{%
100							82.53\\
75							73.11\\
50							74.50\\
25							85.16\\
0							89.03\\
};
\addlegendentry{Synthetic}

\addplot [color=mycolor1,dotted, mark=*, line width = 0.7mm]
table[row sep=crcr]{%
	100														70.48\\
	75														62.44\\
	50														67.39\\
	25														69.55\\
	0														100.00\\
};
\addlegendentry{Synthetic, FT}

%\addplot [color=mycolor1,dashed, mark=*, line width = 0.7mm]
%table[row sep=crcr]{%
%	100																					74.81\\
%	75																					58.58\\
%	50																					67.08\\
%	25																					74.19\\
%	0																					100.00\\
%};
%\addlegendentry{Synthetic, FT 3Ep}

\addplot [color=mycolor2, mark=triangle, line width = 0.7mm]
  table[row sep=crcr]{%
100							47.60\\
75							53.79\\
50							50.70\\
25							55.49\\
0							80.22\\
};
\addlegendentry{Partly-real}

\addplot [color=mycolor2,dotted, mark=triangle, line width = 0.7mm]
table[row sep=crcr]{%
	100														41.89\\
	75														47.14\\
	50														41.42\\
	25														56.11\\
	0														100.00\\
};
\addlegendentry{Partly-real, FT}

%\addplot [color=mycolor2,dashed, mark=triangle, line width = 0.7mm]
%table[row sep=crcr]{%
%	100																					44.67\\
%	75																					59.35\\
%	50																					52.24\\
%	25																					71.10\\
%	0																					100.00\\
%};
%\addlegendentry{PR, FT 3Ep}

%\addplot [color=mycolor3, mark=*, dash pattern=on 1pt off 3pt on 3pt off 3pt, line width = 0.7mm]
\addplot [color=mycolor3, mark=diamond, line width = 0.7mm]
  table[row sep=crcr]{%
100							47.30\\
75							41.27\\
50							48.69\\
25							46.21\\
0							68.16\\
};
\addlegendentry{Synthetic+Partly-real}

\addplot [color=mycolor3, dotted, mark=diamond, line width = 0.7mm]
table[row sep=crcr]{%
	100														39.88\\
	75														50.23\\
	50														41.58\\
	25														42.16\\
	0														94.59\\
};
\addlegendentry{Synthetic+Partly-real, FT}

%\addplot [color=mycolor3, dashed, mark=diamond, line width = 0.7mm]
%table[row sep=crcr]{%
%	100																					43.12\\
%	75																					52.40\\
%	50																					50.85\\
%	25																					47.30\\
%	0																					100.00\\
%};
%\addlegendentry{Syn+PR, FT 3Ep}

%\addplot [color=mycolor4, mark=*, dash pattern=on 3pt off 6pt on 6pt off 6pt, line width = 0.7mm]
\addplot [color=mycolor4, line width = 0.7mm] %mark=square,
  table[row sep=crcr]{%
100	100 \\
75	100 \\
50	100 \\
25	100 \\
0	100 \\
};
\addlegendentry{Baseline (real only)}

\end{axis}
\end{tikzpicture}%
}
\caption{MR}
\label{fig:MR}
\end{subfigure}
\caption{The plots show the results in terms of CER and MR for the algorithm trained on the data sets of different size and type. The algorithm trained on the synthetic, partly-real and the combined training sets are denoted in blue, orange and yellow respectively. The solid lines denote the algorithm trained on the artificial training data only. The finetuned versions of the respective model are denoted by dotted lines of the same color. Additionally, the baseline performance is marked in purple.}
\label{fig:results}
\end{figure*}
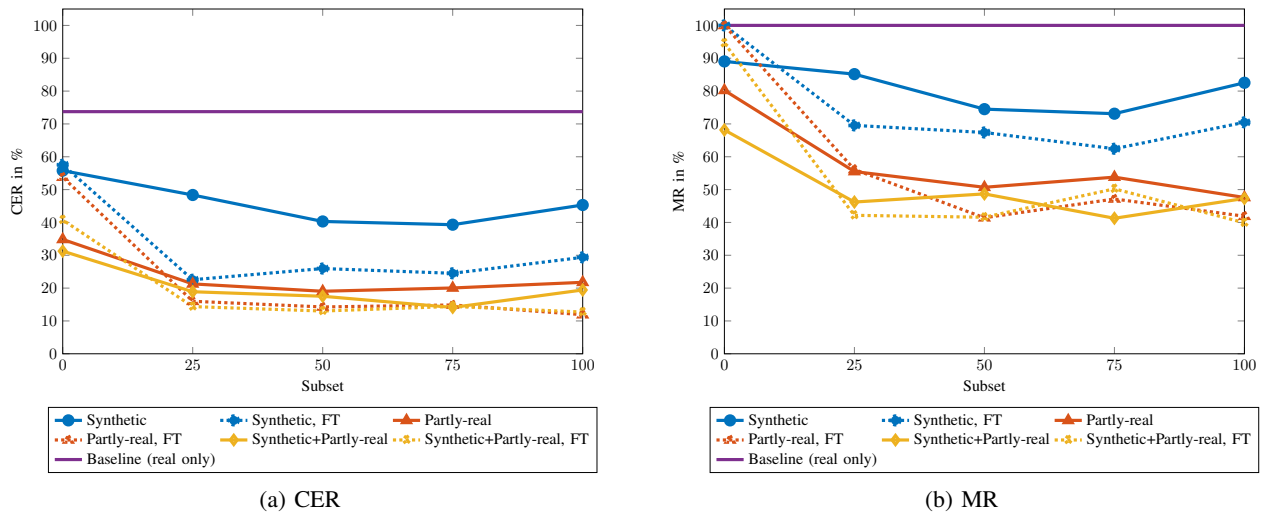
All above mentioned results are summarized and depicted in Figure~\ref{fig:results}. The solid lines denote the results of the algorithm trained on the artificial training data only. Here the results of the algorithm trained on the synthetic, partly-real and the combined training sets are plotted in blue, orange and yellow respectively. The finetuned versions of the respective model are plotted by dotted lines of the same color. Additionally, the baseline performance of the algorithm trained on the real-world data set solely is marked in purple. In Figure~\ref{fig:CER} the CER is plotted over the training subsets of different size. Analogously, in Figure~\ref{fig:MR} the MR is plotted. The plots in Figure~\ref{fig:results} mark the large gains in terms of CER as well as MR enabled by using training data synthesized with our proposed rendering-based data generation pipeline.\par 
Analyzing the plots in Figure~\ref{fig:results}, it can be observed that the error rates rise again for training on larger subsets, especially for training on synthetic data solely. This can be explained by beginning overfitting onto the synthetic data. By a subsequent finetuning on real data the effects of the beginning overfitting can be alleviated and it may even improve the overall performance as shown especially for the combination of synthetic and partly-real data sets. The required amount of training samples depends on the deployed algorithm and considered recognition task. As the proposed process of synthesizing synthetic and partly-real data is entirely automated, the required amount of training data can be generated by the user matching the individual setup. Additionally, Figure~\ref{fig:results} shows the improved performance of the algorithm by incorporating partly-real data during the training process, as the error rates can be decreased significantly. The partly-real data introduce real-world camera artifacts in the training and can thereby improve the final performance on real test data. \par
In the application examined by the experiments, our data generation pipeline enabled the usage of Tesseract for forensic ALPR tasks. The Tesseract version, trained on real training data solely, is not applicable for ALPR applications due to its very high error rates of $73.74\%$ in terms of CER and $100\%$ in terms of MR. By additionally using the training data generated using our entirely automatized rendering-based data generation pipeline the error rate could be reduced significantly to $12.71\%$ in terms of CER and $39.88\%$ in terms of MR. This marks a clearly improved recognition of the real-world license plate samples, especially for the forensic application of license plate recognition considered here. The deployed test set holds a notable number of very challenging samples with small resolution, strong blur, bad illumination and adverse viewing angles, as for example shown in the bottom line of Figure~\ref{fig:real_LP}.
\section{Conclusion}
\label{sec:conclusion}
In this paper we proposed a novel rendering-based pipeline for synthesizing image and video data. Our proposed pipeline generates realistic samples by incorporating light sources producing physically correct illumination scenarios. Besides the generation of synthetic data the pipeline also aids the acquisition of real data. A further contribution of the proposed pipeline is a framework for the generation of partly-real data. The partly-real data is able to narrow down the gap between the synthetic and real data, by incorporating real-world cameras and lenses. The generation and annotation process of the synthetic and partly-real data is entirely automated. This allows an easy generation of large-scale data sets.\par 
Within the conducted experiments, we could demonstrate the advantages of incorporating training data produced by our proposed data generation pipeline. The proposed data generation pipeline is especially advantageous for scenarios with only very limited amount of available training samples. Here we examined the effect of incorporating artificial training data for an forensic ALPR application. By additionally using the data generated with the proposed data generation pipeline, we were able to show a reduction of the CER from $73.74\%$ to $11.90\%$. Also the MR could be decreased significantly from $100\%$ to $39.88$. \par
A significant reduction of the recognition error rates could be achieved without incorporating any real-world data during training, but only using synthetic and partly-real data that was generated and annotated by our proposed pipeline in a fully automated manner. The CER and MR could be reduced from $73.74\%$ and $100\%$ to $14.11\%$ and $41.27\%$ respectively by training on artificial data solely. \par
Thereby, the proposed rendering-based pipeline for the automated generation of image and video data is capable of significantly decreasing the initial overhead of deploying a machine learning algorithm for applications where none or only limited annotated data sets are available. The proposed rendering-based data generation pipeline including all code, as well as the data sets used for the conducted experiments are made publicly available under~\textit{(URL will be revealed upon publication)}.

\section*{Acknowledgments}
%
%We gratefully acknowledge support by the German Federal Ministry of Education and Research (BMBF) under Grant No.~13N15319. \par  
We would like to thank Abinaya Aravindan, Aalok Gutpa, and Nishant Jaiswal for their help with the recordings.
\bibliographystyle{IEEEtran}
\bibliography{bib_strings_long,bibliography}
\begin{IEEEbiography}[{\includegraphics[width=1in,height=1.25in,clip,keepaspectratio]{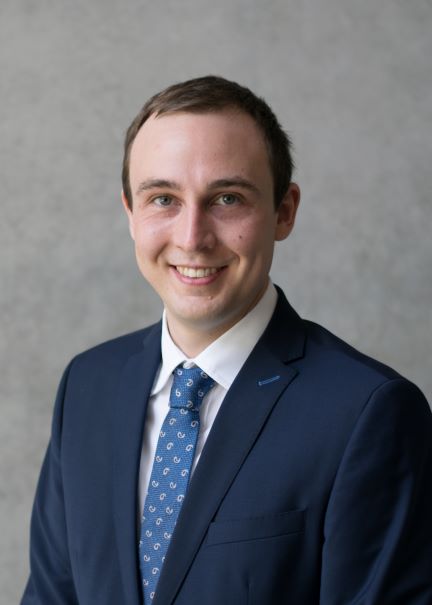}}]{Andreas Spruck}
received his B.Sc. and M.Sc. degree in electrical engineering and information technology from the Friedrich-Alexander-Universit\"at Erlangen-N\"urnberg (FAU), Germany, in 2015 and 2017, respectively. He is currently pursuing the Dr.-Ing. degree with the Chair of Multimedia Communications and Signal Processing, FAU Erlangen-N\"urnberg. His research interests include the processing of image and video data with a special focus on the influence of these methods onto the performance of NNs. 
\end{IEEEbiography}
\begin{IEEEbiography}[{\includegraphics[width=1in,height=1.25in,clip,keepaspectratio]{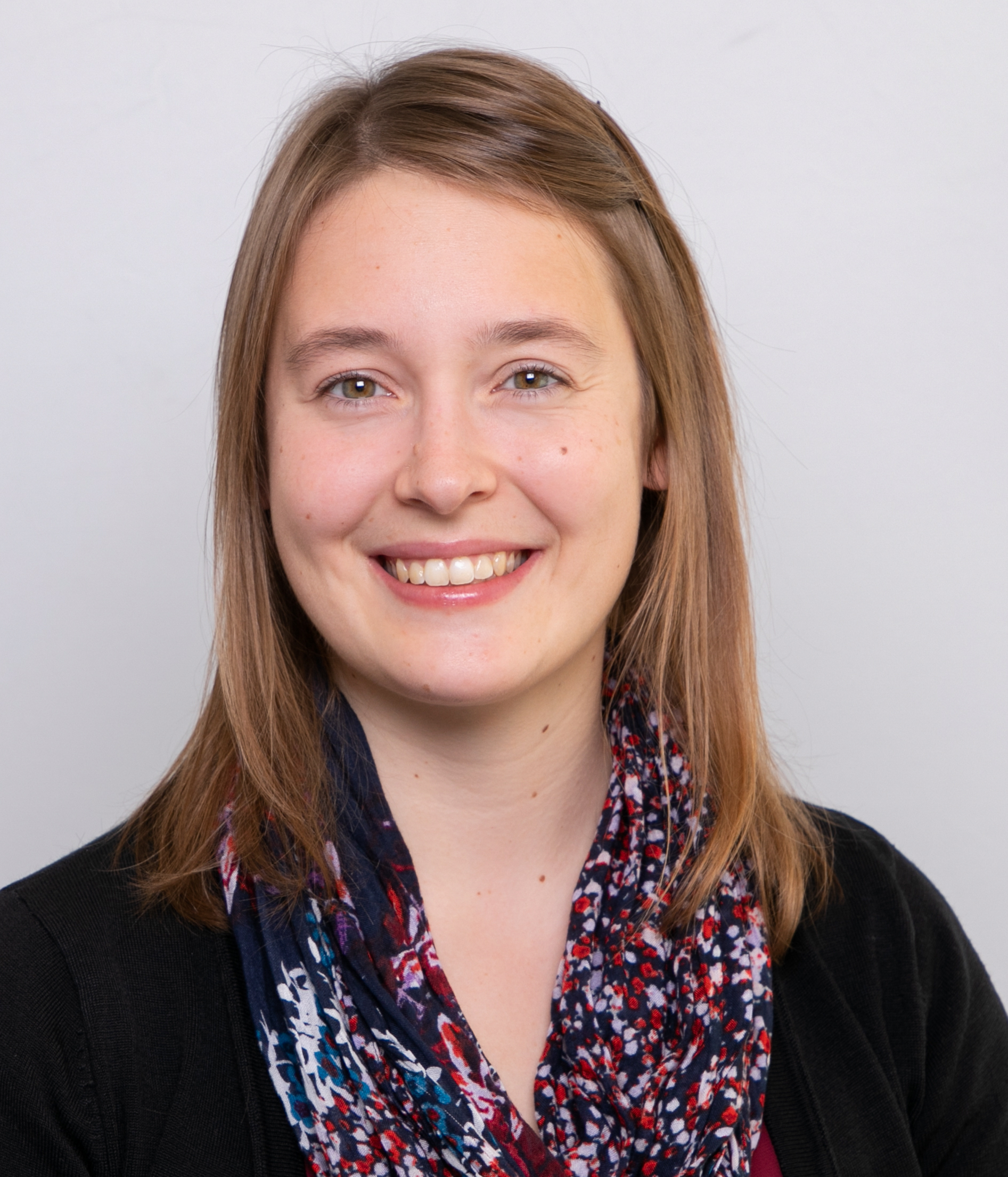}}]{Maximiliane Gruber}
	received her B.Sc. degree in medical engineering and her M.Sc. degree in information and communication technology from Friedrich-Alexander-Universit\"at Erlangen-N\"urnberg (FAU), Germany, in 2017 and 2020, respectively. Since 2020, she has been a Researcher with the Chair of Multimedia Communications and Signal Processing at FAU, where she conducts research on image and video signal processing. Her current research interests include pixel-level cross-camera domain adaptation.
\end{IEEEbiography}
\begin{IEEEbiography}[{\includegraphics[width=1in,height=1.25in,clip,keepaspectratio]{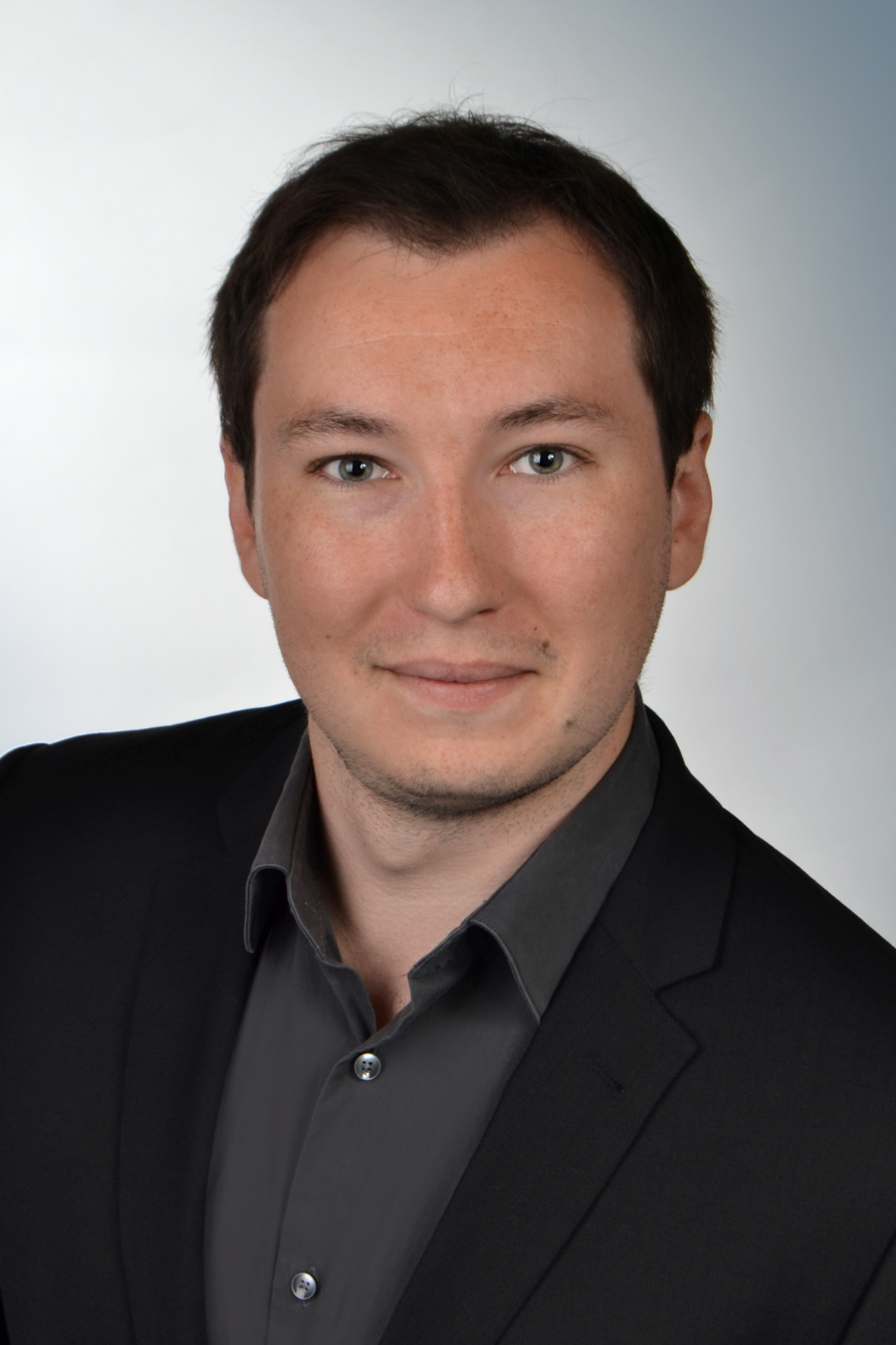}}]{Anatol Maier}
	received the M.Sc. degree in computer science from the Friedrich-Alexander-Universit\"at Erlangen-N\"urnberg (FAU), Germany, in 2019. Since November 2019, he is Ph.D. student at the IT Security Infrastructures Lab at FAU	and part of the Multimedia Security Group. His research interests include reliable machine learning, deep probabilistic models, and computer vision with particular application in image and video forensics.	
\end{IEEEbiography}
\begin{IEEEbiography}[{\includegraphics[width=1in,height=1.25in,clip,keepaspectratio]{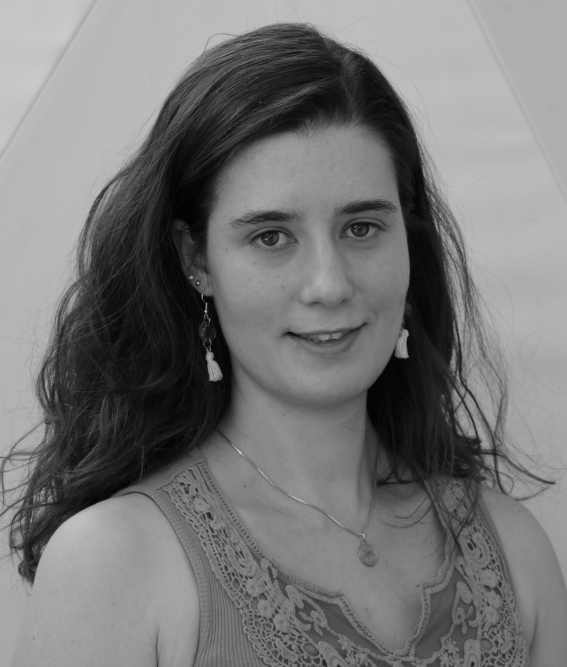}}]{Denise Moussa}
	received the B.Sc degree in computer science in 2018 and the M.Sc  degree in computer science in 2020 from the Friedrich-Alexander-Universit\"at Erlangen-N\"urnberg (FAU), Germany. Since 2021, she is part of the Multimedia Security Group at the IT Security Infrastructures Lab at FAU. Her research interest include computer vision, image and audio forensics, as well as machine learning.
\end{IEEEbiography}
\begin{IEEEbiography}[{\includegraphics[width=1in,height=1.25in,clip,keepaspectratio]{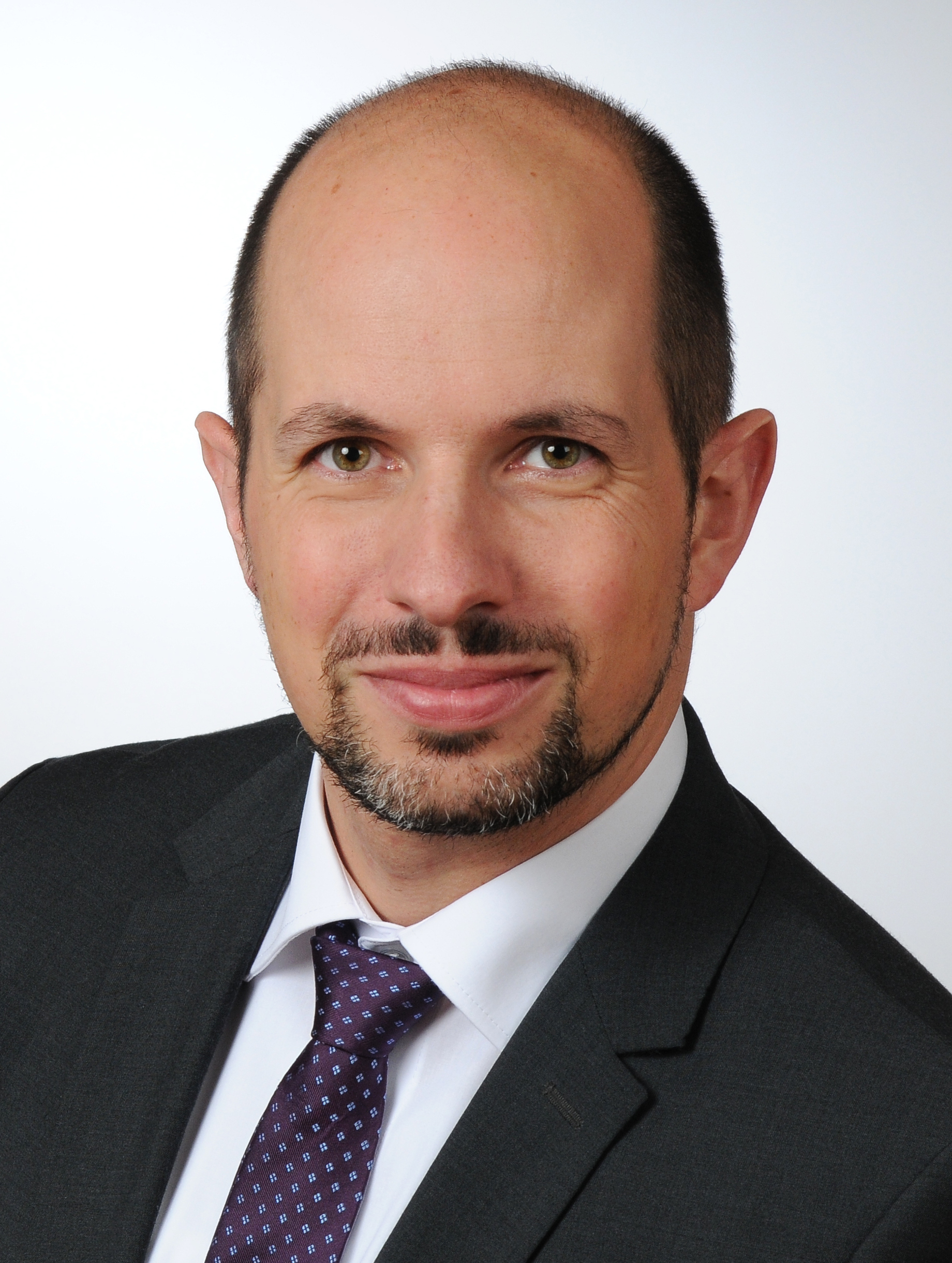}}]{J\"urgen Seiler}
	(M'09-SM'16) is senior scientist and lecturer at the Chair of Multimedia Communications and Signal Processing at the Friedrich-Alexander-Universit{\"a}t Erlangen-N{\"u}rnberg (FAU), Germany. There, he also received his habilitation degree in 2018, his doctoral degree in 2011, and his diploma degree in Electrical Engineering, Electronics and Information Technology in 2006.
	
	He received the dissertation award of the Information Technology Society of the German Electrical Engineering Association as well as the dissertation award of the Staedtler-Foundation, both in 2012. In 2007, he received diploma awards from the Institute of Electrical Engineering, Electronics and Information Technology, Erlangen, as well as from the German Electrical Engineering Association. He also received scholarships from the German National Academic Foundation and the Lucent Technologies Foundation. He is the co-recipient of five best paper awards and he has authored or co-authored more than 100 technical publications.
	
	His research interests include image and video signal processing, signal reconstruction and coding, signal transforms, and linear systems theory.
\end{IEEEbiography}
\begin{IEEEbiography}[{\includegraphics[width=1in,height=1.25in,clip,keepaspectratio]{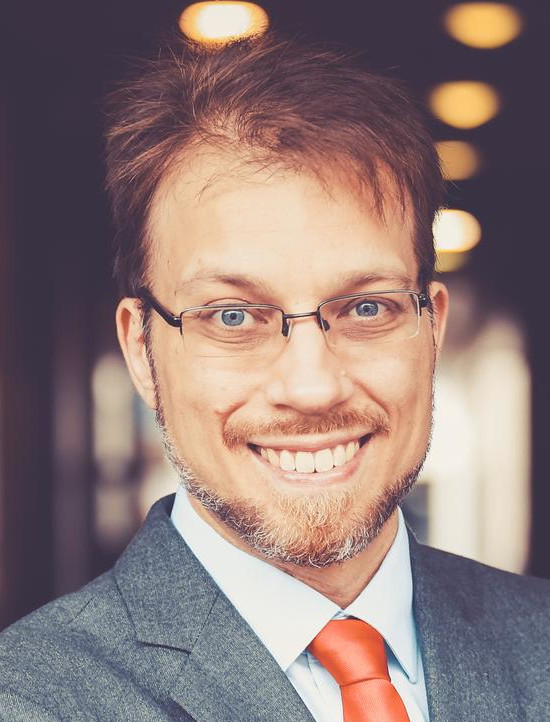}}]{Christian Riess}
	received the Ph.D.\ degree in computer science from Friedrich-Alexander-Universit\"at Erlangen-N\"urnberg (FAU), Germany, in 2012. From 2013 to 2015, he was a Postdoc at the Radiological Sciences Laboratory at Stanford University. Since 2016, he is senior researcher and head of the Multimedia Security Group at the IT Infrastructures Lab at FAU. He received his habilitation on Phase-Contrast X-ray Imaging from FAU in 2020.
	
	Dr.\ Riess is currently a member of the IEEE Information Forensics and Security Technical Committee and the EURASIP TAC Signal and Data Analysis. His research interests include all aspects of image processing and imaging, particularly with applications in image and video forensics, X-ray phase contrast, color image processing, and computer vision. His work on copy-move forgery detection received the 2017 IEEE Signal Processing Society Award.
\end{IEEEbiography}
\begin{IEEEbiography}[{\includegraphics[width=1in,height=1.25in,clip,keepaspectratio]{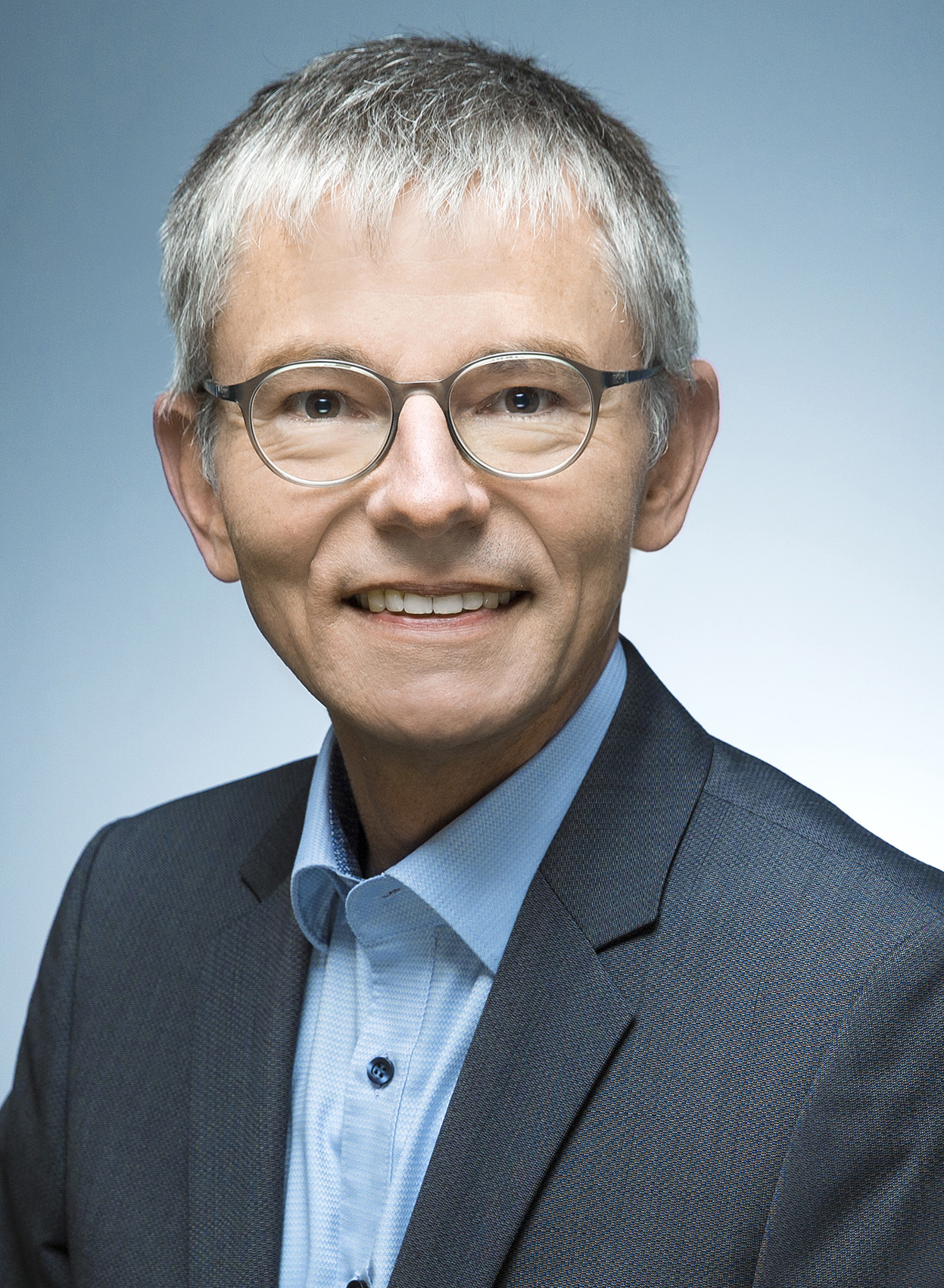}}]{Andr\'e Kaup}
	(Fellow, IEEE) received the Dipl.-Ing. and Dr.-Ing. degrees in electrical engineering from RWTH Aachen University, Aachen, Germany, in 1989 and 1995, respectively.
	
	He joined Siemens Corporate Technology, Munich, Germany, in 1995, and became the Head of the Mobile Applications and Services Group in 1999. Since 2001, he has been a Full Professor and the Head of the Chair of Multimedia Communications and Signal Processing at Friedrich-Alexander-Universit\"at Erlangen-N\"urnberg (FAU), Germany. From 2005 to 2007 he was Vice Speaker of the DFG Collaborative Research Center 603. From 2015 to 2017, he served as the Head of the Department of Electrical Engineering and Vice Dean of the Faculty of Engineering at FAU. He has authored around 400 journal and conference papers and has over 120 patents granted or pending. His research interests include image and video signal processing and coding, and multimedia communication.
	
	Dr. Kaup is a member of the Scientific Advisory Board of the German VDE/ITG. He served as a member of the IEEE Multimedia Signal Processing Technical Committee. He is an IEEE Fellow and a member of the Bavarian Academy of Sciences. He is a member of the Editorial Board of the IEEE Circuits and Systems Magazine. He was a Siemens Inventor of the Year 1998 and obtained the 1999 ITG Award. He received several IEEE best paper awards, including the Paul Dan Cristea Special Award in 2013, and his group won the Grand Video Compression Challenge from the Picture Coding Symposium 2013. The Faculty of Engineering with FAU and the State of Bavaria honored him with Teaching Awards, in 2015 and 2020, respectively. He served as an Associate Editor of the IEEE Transactions on Circuits and Systems for Video Technology. He was a Guest Editor of the IEEE Journal of Selected Topics in Signal Processing.
\end{IEEEbiography}
\vfill
\end{document}